\documentclass[10pt,twocolumn,letterpaper]{article}

\usepackage{cvpr}              

\usepackage{graphicx}
\usepackage{subcaption}
\usepackage{float}
\usepackage{algorithm}
\usepackage{algpseudocode}
\usepackage{amsmath}
\usepackage{mathtools}
\usepackage[table]{xcolor}
\usepackage[most]{tcolorbox}
\usepackage{booktabs}       
\usepackage{multirow}    
\usepackage{siunitx} 
\usepackage{array}
\usepackage{caption}
\usepackage{geometry}
\definecolor{avgbg}{HTML}{F7E3C8}    
\definecolor{inc}{HTML}{0F9D58}      
\definecolor{dec}{HTML}{D93025}      
\definecolor{lightgray}{gray}{0.95}  
\newcommand{\plus}[1]{\textsuperscript{\textcolor{inc}{+#1}}}
\newcommand{\minusv}[1]{\textsuperscript{\textcolor{dec}{-#1}}}
\newtheorem{theorem}{Theorem}
\definecolor{cvprblue}{rgb}{0.21,0.49,0.74}
\usepackage[pagebackref,breaklinks,colorlinks,allcolors=cvprblue]{hyperref}

\def\paperID{144} 
\def\confName{CVPR}
\def\confYear{2026}

\title{SOTA: Self-adaptive Optimal Transport for Zero-Shot Classification with Multiple Foundation Models}

\author{
Zhanxuan Hu$^{1}$ \quad Qiyu Xu$^{3}$ \quad Yu Duan$^{2}$$^{\dagger}$ \quad Yonghang Tai$^{1}$$^{\dagger}$ \quad Huafeng Li$^{4}$\\
$^{1}$Yunnan Normal University, 
$^{2}$Xidian University \\
$^{3}$Xi’an University of Posts and Telecommunications \\
$^{4}$Kunming University of Science and Technology \\
\tt\small{ \{zhanxuanhu, graceafleve, duanyuee\}@gmail.com }\\
\tt\small{taiyonghang@126.com}, \tt\small{hfchina99@163.com} \\ 
}

\begin{document}
\maketitle

\begin{abstract}

Foundation models have attracted widespread attention across domains due to their powerful zero-shot classification capabilities. This work is motivated by two key observations: (1) \textit{Vision-Language Models} (VLMs), such as CLIP, often over-rely on class-level textual priors and struggle to capture fine-grained visual cues, whereas \textit{Vision-only Foundation Models} (VFMs), such as DINO, provide rich and discriminative visual features but lack semantic alignment; (2) the performance of different VLMs varies considerably across datasets owing to differences in pre-training. To address these challenges, we propose \textbf{SOTA} (\textit{Self-adaptive Optimal TrAnsport}), a \textit{training-free} ensemble framework that integrates the outputs of multiple foundation models~(VFMs or VLMs) by learning a self-adaptive transport plan. Notably, \textbf{SOTA} is prior-free and automatically balances model contributions. Extensive experiments across diverse domains, including natural images, medical pathology, and remote sensing, validate the generalizability of \textbf{SOTA}. The results consistently show that it effectively leverages the complementary strengths of different foundation models and achieves substantial improvements over individual models. The implementation code is available at: \url{https://github.com/Afleve/self-adaptive-Optimal-Transport}.



\end{abstract}
    
\section{Introduction}

\renewcommand{\thefootnote}{\fnsymbol{footnote}} 
\footnotetext[2]{Corresponding Authors}

By leveraging large-scale pre-training, foundation models such as CLIP~\cite{radford2021learning} and DINO~\cite{oquab2023dinov2,dinov3} learn generalizable representations that can be directly applied to zero-shot classification without additional supervision. 
Thanks to their remarkable zero-shot transfer capabilities, foundation models have received growing attention from the research community. Consequently, various models have been developed and successfully applied across diverse domains, including natural images~\cite{siglip}, medical pathology~\cite{conchlu2024visual,muskxiang2025vision}, and remote sensing~\cite{georsclip,skyscript}.

\begin{figure}[t]
    \centering
    \includegraphics[width=0.45\textwidth]{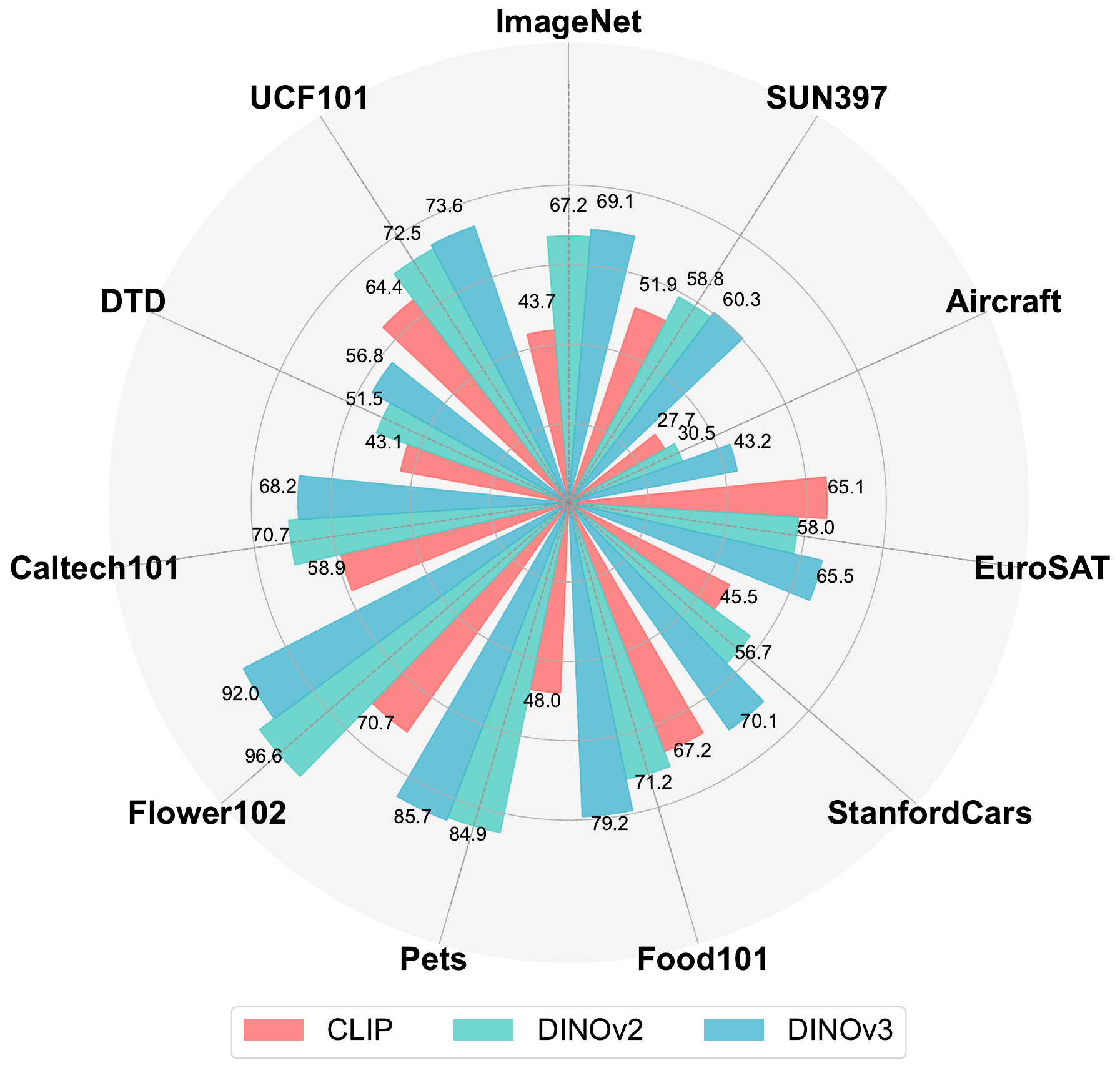}
    \caption{Clustering accuracy comparison of visual features extracted from different foundation models. Compared with VLMs, VFMs produce more discriminative representations, especially on fine-grained datasets such as \texttt{StanfordCars}, \texttt{Flower102}, and \texttt{Pets}.}
    \vspace{-0.5cm}
    \label{Fig:clip_v1_v2}
\end{figure}

Beyond designing new foundation models, another important line of research focuses on improving the zero-shot transfer ability of existing ones. Representative approaches include prompt engineering~\cite{vlm-1, vlm-2}, label propagation~\cite{kalantidis2024label,li2025efficient}, and distribution alignment~\cite{transductive-3,transductive-4}. While these methods have achieved encouraging results, they typically focus on enhancing a \textit{single} model and overlook an important perspective: \emph{due to differences in pre-training settings, foundation models often possess distinct characteristics.}
\begin{figure*}[t]
  \centering
  \begin{subfigure}[b]{0.305\textwidth}
    \centering
    \includegraphics[width=\linewidth]{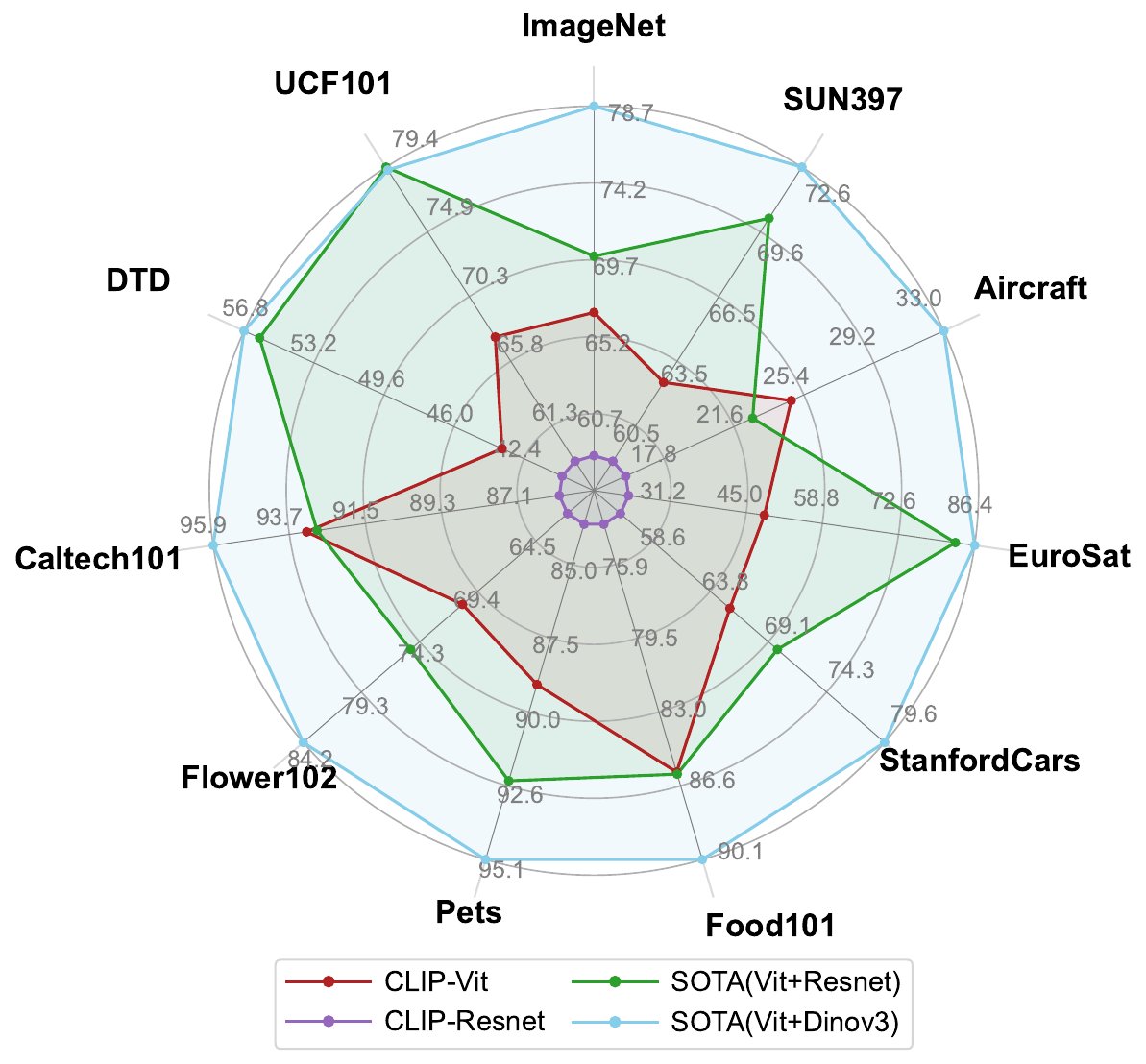}
    \caption{Natural images}
    \label{fig:sub:nature}
  \end{subfigure}\hfill
  \begin{subfigure}[b]{0.32\textwidth}
    \centering
    \includegraphics[width=\linewidth]{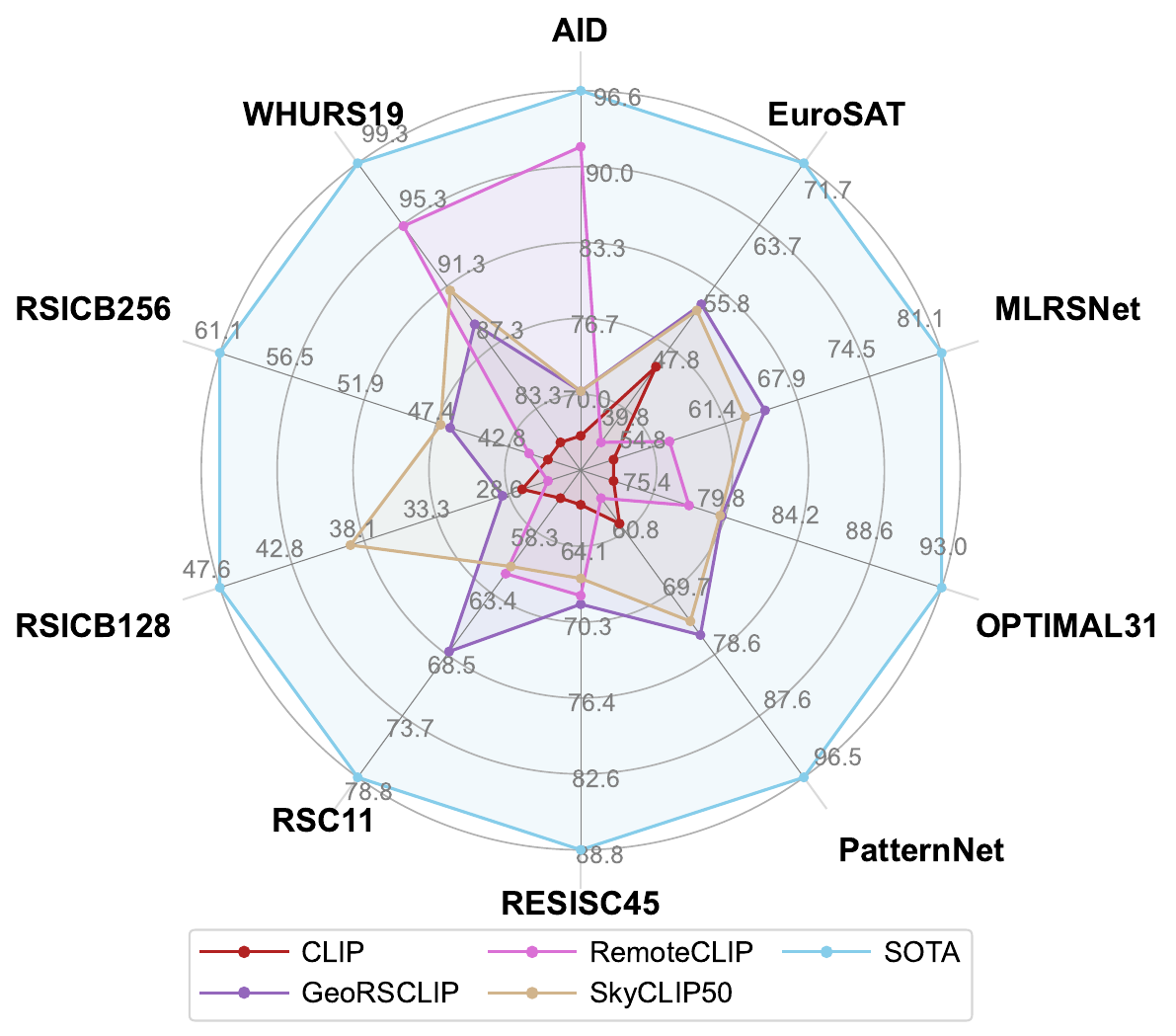}
    \caption{Remote sensing}
    \label{fig:sub:sensing}
  \end{subfigure}\hfill
  \begin{subfigure}[b]{0.335\textwidth}
    \centering
    \includegraphics[width=\linewidth]{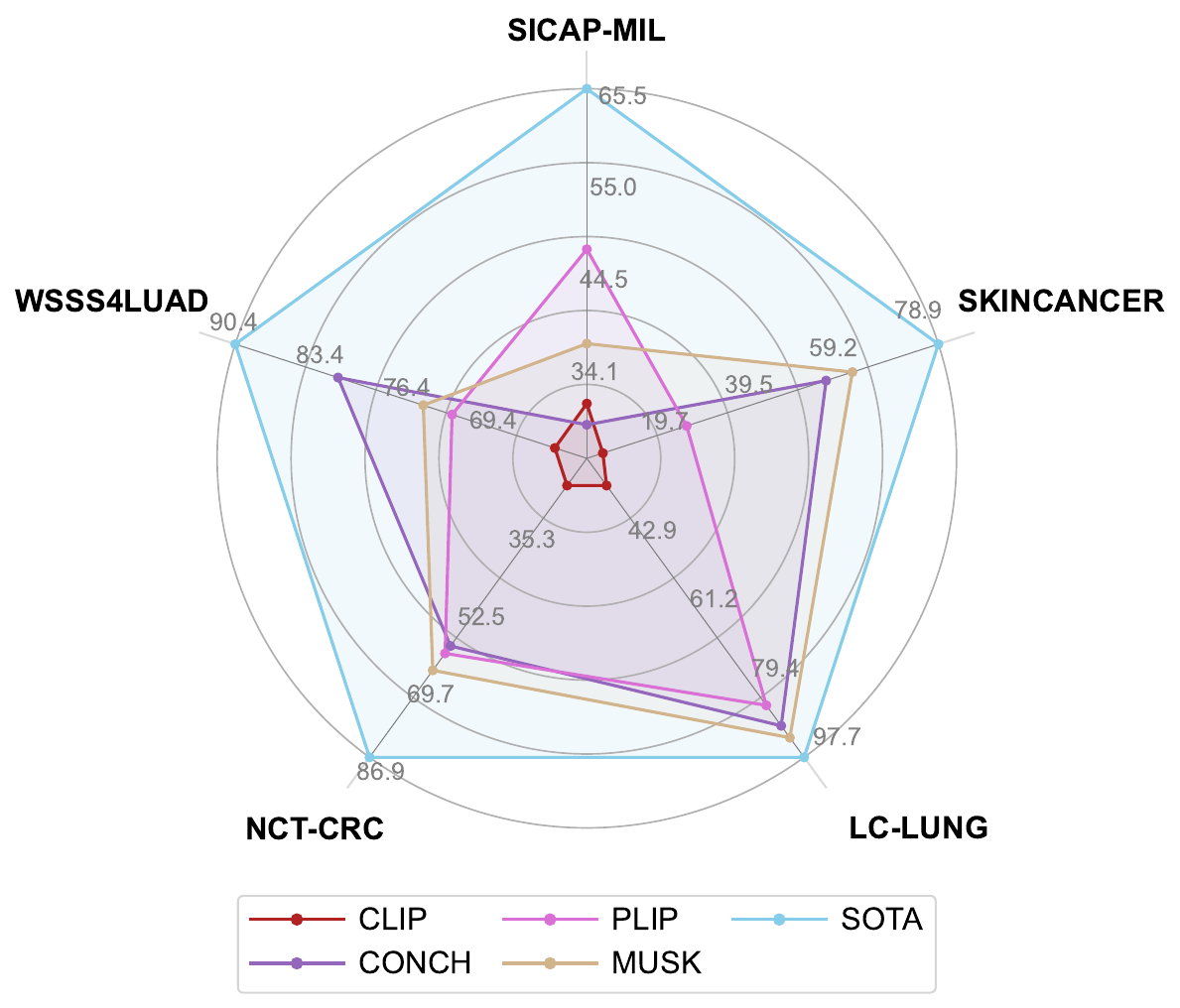}
    \caption{Medical pathology}
    \label{fig:sub: medicine}
  \end{subfigure}
\caption{Top-1 accuracy of \textbf{SOTA} in zero-shot classification across three domains. The performance of individual VLMs varies notably across datasets, while \textbf{SOTA} effectively exploits their complementary strengths, yielding substantial improvements.}
\vspace{-0.5cm}
  \label{Fig:ks}
\end{figure*}

This work is inspired by two key observations: First, \textit{Vision-Language Models} (VLMs), such as CLIP~\cite{radford2021learning}, possess strong cross-modal alignment capabilities that enable impressive zero-shot transfer. However, their visual encoders tend to over-rely on class-level textual priors and often fail to capture fine-grained visual cues~(see Figure~\ref{Fig:clip_v1_v2}), which are crucial for distinguishing visually similar categories. In contrast, \textit{Vision-only Foundation Models} (VFMs), such as DINO~\cite{oquab2023dinov2,dinov3}, provide rich and discriminative visual representations, but inherently lack semantic alignment with category labels. Second, the performance of different VLMs varies considerably across datasets due to differences in pre-training~(see Figure~\ref{Fig:ks}). These observations naturally raise an important question: 

\begin{tcolorbox}[colback=cyan!10, colframe=cyan!10, boxsep=0pt, arc=0mm]
\emph{Rather than relying solely on a single foundation model, can we enhance zero-shot classification by integrating multiple foundation models to fully exploit their complementary strengths?}
\end{tcolorbox}

To this end, we propose a simple yet effective foundation model integration framework, namely \textbf{SOTA} (\textbf{S}elf-adaptive \textbf{O}ptimal \textbf{T}r\textbf{A}nsport). 
The key idea of \textbf{SOTA} is to regard each foundation model as a distinct view for measuring the relevance between samples and candidate classes. In practice, this relevance is represented by a cost matrix. Instead of heuristically selecting or manually weighting these cost matrices, \textbf{SOTA} introduces a \textit{self-adaptive optimal transport} mechanism that jointly considers all cost matrices to learn a transport plan, a soft assignment from samples to classes, that minimizes the overall transport cost. In the transductive setting, the transport plan is directly used as the final prediction. In the inductive setting, it serves as supervisory guidance, enabling different models to learn individual classifiers, which are then combined during inference to produce the final prediction for the test data. 

\textbf{SOTA} is \textit{training-free} and \textit{prior-free}, automatically adapting to dataset-specific characteristics and avoiding over-reliance on any single model. Extensive experiments across diverse domains, including natural images, medical pathology, and remote sensing, demonstrate the generalizability of our approach. 
As shown in Fig.~\ref{Fig:ks}, \textbf{SOTA} effectively integrates the complementary strengths of different foundation models, achieving substantial improvements over single models. The main contributions of this work are summarized as follows:
\begin{itemize}
    \item \textbf{Novel perspective.} To the best of our knowledge, this is the first work to systematically investigate the complementary strengths of different foundation models in zero-shot classification, opening a new direction for improving zero-shot transfer ability through multi-model integration.
    \item \textbf{Novel method.} We propose a simple yet effective foundation model integration framework \textbf{SOTA}. Notably, \textbf{SOTA} does not require access to the weights of foundation models and can thus enhance the zero-shot performance of black-box models, even when they are accessible solely through an API.
    \item \textbf{Promising results.} We validate \textbf{SOTA} on 26 benchmarks spanning natural, medical pathology, and remote sensing domains, achieving substantial accuracy gains over the best single model, without requiring fine-tuning or additional supervision.
\end{itemize}

\section{Related Work}
\subsection{Foundation models}
Foundation models (FMs) refer to large-scale pre-trained models that are trained on massive datasets and can be adapted to a wide range of downstream tasks with little or no task-specific supervision. They have reshaped the paradigm of computer vision and natural language processing, offering powerful general-purpose representations that significantly reduce the reliance on annotated data. Broadly, existing FMs can be categorized into two groups: \textit{Vision-only Foundation Models (VFMs)} and \textit{Vision-Language Models (VLMs)}.
\paragraph{Vision-only Foundation Models (VFMs).}  
VFMs, such as DINO~\cite{oquab2023dinov2,dinov3}, focus purely on visual pre-training. 
They excel at learning generic and transferable representations from large-scale unlabeled image collections, which makes them highly effective in open-set or cross-domain scenarios. 
However, the absence of semantic alignment with textual information limits their ability to directly handle tasks that involve fine-grained category definitions, such as zero-shot classification and open-vocabulary semantic segmentation.
\paragraph{Vision-Language Models (VLMs).}  
VLMs, exemplified by CLIP~\cite{radford2021learning}, learn to align visual and textual modalities through contrastive pre-training on large-scale image-text pairs. Such multimodal alignment allows VLMs to directly associate an unseen class name with its corresponding visual instances, thereby making them naturally suitable for zero-shot classification. However, their reliance on noisy web-scale image-text pairs may introduce semantic bias, and the performance often degrades in domain-specific scenarios (e.g., medical imaging or remote sensing), where the pre-training distribution substantially differs from the target domain. 
To alleviate this limitation, a number of domain-adapted VLMs have been developed. Representative examples include {CONCH}~\cite{conchlu2024visual}, {PLIP}~\cite{pliphuang2023visual}, and {MUSK}~\cite{muskxiang2025vision} for medical pathology, as well as GeoRSCLIP~\cite{georsclip}, RemoteCLIP~\cite{remoteclip}, and SkyCLIP~\cite{skyscript} for remote sensing.
\subsection{VLM-based Zero-shot Classification}
Despite their impressive generalization ability, VLMs still face several limitations when applied to zero-shot classification. 
First, they often struggle to capture fine-grained visual features, which are critical for distinguishing visually similar categories~\cite{huang2025cosmic}. 
Second, the inherent modality gap between visual and textual representations may hinder accurate alignment~\cite{qian2023intra, huang2025enhance}. 
To address these issues, various approaches have been proposed, including vision fine-tuning~\cite{lafter, pouf}, text prompt engineering~\cite{CuPL,TPT,Promptkd,vlm-1, vlm-2,udandarao2023sus}, visual modality classifier learning~\cite{GDA, qian2023intra,zhu2025dynamic}, and test-time adaptation~\cite{huang2025cosmic,karmanov2024efficient}. However, most of these methods require substantial training time and computational resources. 
In contrast, training-free techniques, such as label propagation~\cite{kalantidis2024label,li2025efficient}, leverage the graph structure of unlabeled data to achieve efficient inductive/transductive inference. 
Notably, this work is partly inspired by \cite{transductive-4}, which primarily focuses on improving CLIP’s performance in the transductive setting. 
Different from their approach, our work aims to propose a self-adaptive ensemble framework that integrates the complementary strengths of diverse foundation models.

\subsection{Optimal Transport}
Optimal Transport~\cite{OT} offers a principled framework for comparing and aligning probability distributions by explicitly modeling the geometry of the underlying metric space. In recent years, it has been reintroduced to the machine learning community and applied to a variety of tasks. For instance, Chang et al.~\cite{chang2022unified} leverage OT for domain adaptation, enabling simultaneous common class detection and private class discovery. Nicolas et al.~\cite{chenplot} utilize OT to achieve fine-grained alignment between local visual features and multiple textual prompts, thereby enhancing cross-modal correspondence. Tan et al.~\cite{tan2025recover} reformulate region-to-label matching as an optimal transport problem, significantly boosting multi-label classification performance. In addition, OT has been explored for adapting foundation models~\cite{zhu2025dynamic,awt} and conformal prediction~\cite{OT-CP}.
In contrast to the above works, we employ OT to integrate the outputs of multiple foundation models within a unified framework. This enables more robust and discriminative predictions, without requiring model retraining.


\section{Method}
\begin{figure*}[t]
    \centering
    \includegraphics[width=0.95\textwidth]{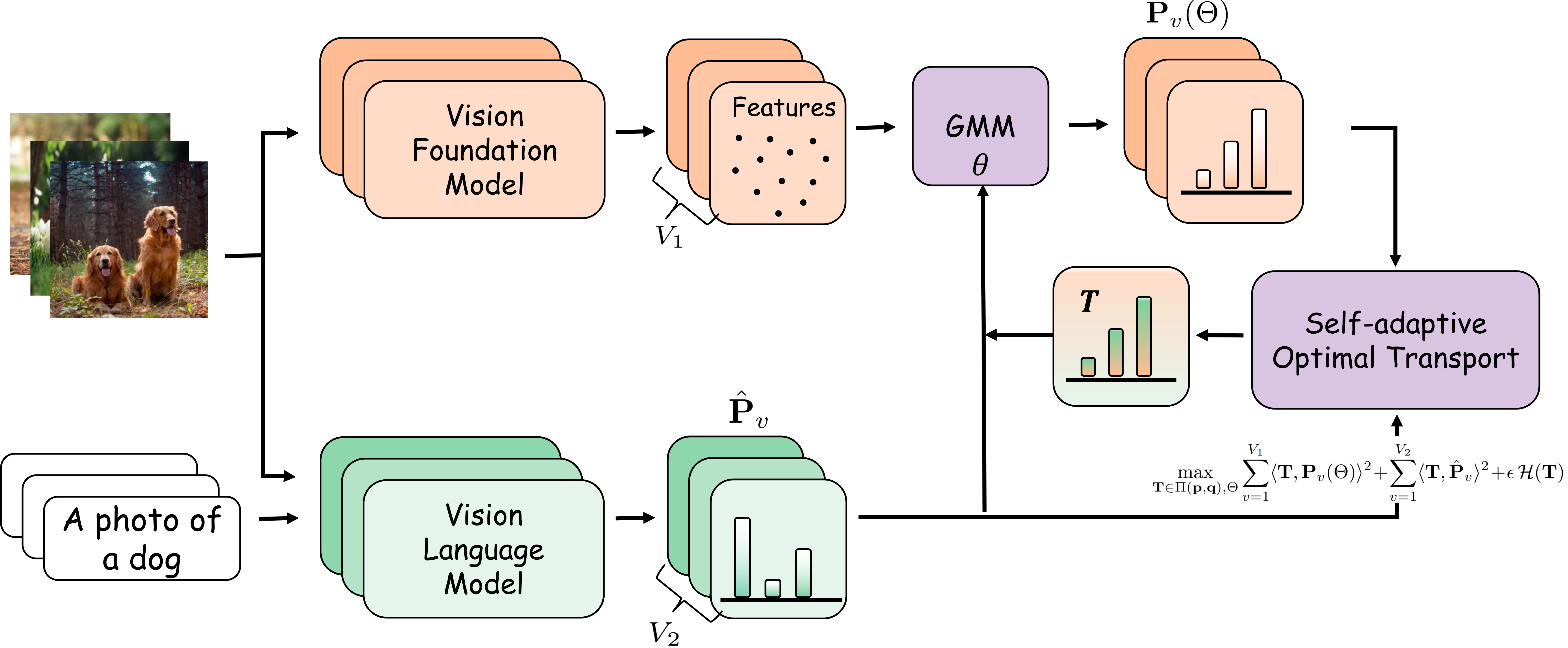}
    \caption{\textbf{Pipeline of our method SOTA}. We adopt a self-adaptive optimal transport strategy to integrate the outputs of diverse foundation models (VFMs or VLMs), yielding a transport plan $\mathbf{T}$. In the transductive setting, $\mathbf{T}$ directly serves as the final prediction. In the inductive setting, $\mathbf{T}$ guides the estimation of GMM parameters $\Theta$, which form one or multiple visual classifiers that collaborate with the text classifier to produce predictions on unseen test data.}
    \vspace{-0.5cm}
    \label{Fig:framework}
\end{figure*}

\subsection{Preliminary}
\paragraph{Problem Setting.} 
Zero-shot classification aims to recognize instances from unseen categories without requiring labeled training samples. 
Formally, we are given an unlabeled dataset $\mathcal{D}_u=\{\mathbf{x}_i\}_{i=1}^N$ and a set of category names or descriptions that serve as side information.  In the \emph{Inductive inference} setting,  $\mathcal{D}_u$ is regarded as training data. 
The goal is to learn a classifier that generalizes to unseen test instances from the same distribution.  In the \emph{Transductive inference} setting, $\mathcal{D}_u$ itself constitutes the entire test set. In this case, the model can exploit the global distributional structure of $\mathcal{D}_u$ to directly infer category assignments, without explicitly learning a separate classifier. In this work, we assume access to multiple types of foundation models, including both VFMs and VLMs~\footnote{It is worth noting that the visual encoder of each VLM can also be regarded as a VFM.}, and aim to integrate their complementary strengths, achieving improved prediction accuracy.
\paragraph{Optimal Transport.}  
OT~\cite{OT} provides a principled mathematical framework for comparing and aligning probability distributions by explicitly modeling the geometry of the underlying metric space. Specifically, given two discrete probability distributions \(\mathbf{p} \in \Delta^N\) and \(\mathbf{q} \in \Delta^K\), OT seeks a transport plan \(\mathbf{T} \in \mathbb{R}^{N \times K}\) that maps mass from \(\mathbf{p}\) to \(\mathbf{q}\) at minimum cost. Formally, the OT problem is defined as:
\begin{equation}\label{eq:1}
\begin{split}
       & \min_{\mathbf{T} \in \Pi(\mathbf{p}, \mathbf{q})} \langle \mathbf{T}, \mathbf{C} \rangle, \\
         \Pi(\mathbf{p}, \mathbf{q}) & = \left\{ \mathbf{T} \geq 0 \,\middle|\, \mathbf{T} \mathbf{1}  = \mathbf{p},\; \mathbf{T}^\top \mathbf{1} = \mathbf{q} \right\}  ,
\end{split}
\end{equation}
where \(\langle \cdot, \cdot \rangle\) denotes the Frobenius inner product, $\mathbf{p}$ and $\mathbf{q}$ represent the source and target probability distributions, respectively. In our context, \(\mathbf{p}\) represents the distribution over $\mathcal{D}_u$ in the feature space extracted by foundation models, and \(\mathbf{q}\) corresponds to the semantic distribution over candidate classes. Correspondingly, \(\mathbf{C} \in \mathbb{R}^{N \times K}\) is the cost matrix that encodes pairwise semantic dissimilarities between samples and class labels.

As directly solving Eq.~\eqref{eq:1} is computationally challenging, a regularized variant of OT is often used. The key idea is to soften the transport plan $\mathbf{T}$ by encouraging high-entropy solutions, leading to smoother and more computationally tractable updates. Formally, the objective is:
\begin{equation}\label{eq:2-1}
\min_{\mathbf{T} \in \Pi(\mathbf{p}, \mathbf{q})} \langle \mathbf{T}, \mathbf{C} \rangle-\epsilon\mathcal{H}(\mathbf{T}),
\end{equation}
where $\mathcal{H}(\mathbf{T}) = -\sum_{i,j} \mathbf{T}_{ij} \log \mathbf{T}_{ij}$ is the entropy regularization term, and $\epsilon$ is a parameter controlling the strength of the smoothing effect. Eq.~\eqref{eq:2-1} can be efficiently solved by the Sinkhorn-Knopp algorithm~\cite{OT}.

\subsection{Self-adaptive Optimal Transport for Foundation Model Integration}
The proposed self-adaptive model fusion framework is illustrated in Figure~\ref{Fig:framework}. Specifically, we start by deriving a categorical probability distribution \(\mathbf{P} \in \mathbb{R}^{N \times K}\), satisfying \(\mathbf{P}\mathbf{1} = \mathbf{1}\), from the output of each foundation model. Each column of \(\mathbf{P}\) encodes the similarity of samples to the corresponding category. Naturally, the corresponding cost matrix can be defined as $\mathbf{C} = \mathbf{E} - \mathbf{P}$, where \(\mathbf{E} \in \mathbb{R}^{N \times K}\) is an all-ones matrix. This straightforward transformation converts high-confidence predictions into low transport costs, and vice versa. Subsequently, we employ a self-adaptive optimal transport method to learn a transport plan $\mathbf{T}$. Notably, in the \emph{Transductive inference} setting, the transport plan is directly used as the final prediction. In the \emph{Inductive inference} setting, it serves as supervisory guidance, enabling different models to learn individual classifiers, which are then combined during inference to produce the final prediction for the test data. Next, we will describe each part in detail.
\subsubsection{Deriving Probability Distributions}
\paragraph{Deriving probability distributions from VFMs.}  
For VFMs, which lack inherent semantic alignment, we first extract visual features \(\mathbf{v}_i = f_v(x_i)\) for each image \(x_i\). To obtain a probabilistic assignment over classes, we fit a Gaussian Mixture Model (GMM) with parameters \(\Theta = \{\pi_k, \mathbf{\mu}_k, \mathbf{\Sigma}_k\}_{k=1}^K\) to these features, where \(\pi_k\), \(\mathbf{\mu}_k\), and \(\mathbf{\Sigma}_k\) are mixture weights, means, and covariance matrices, respectively. The posterior probability of each class for a given feature is computed as:
\begin{equation}
    p(y = k \mid \mathbf{v}_i) = \frac{\pi_k \mathcal{N}(\mathbf{v}_i \mid \mathbf{\mu}_k, \mathbf{\Sigma}_k)}{\sum_{j=1}^K \pi_j \mathcal{N}(\mathbf{v}_i \mid \mathbf{\mu}_j, \mathbf{\Sigma}_j)},
\label{eq:3}
\end{equation}
where \(\mathcal{N}(\cdot)\) denotes the Gaussian density function. Stacking these posterior probabilities for all samples forms the matrix \(\mathbf{P} \in \mathbb{R}^{N \times K}\), representing the VFM-induced class distributions. Similarly, assuming that we have $V_1$ different VFMs, we obtain $V_1$ probability distributions
$\{{\mathbf{P}}_v\}_{v=1}^{V_1}$. It is worth noting that the parameters $\Theta$ of a GMM, once learned on the training dataset, can naturally induce a classifier, which can then be applied to make predictions on the test data.
\paragraph{Deriving probability distributions from VLMs.}  
Given a VLM with an image encoder \(f_I\) and a text encoder \(f_T\), we first extract image features \(\mathbf{v}_i = f_I(x_i)\) for each \(x_i\) and obtain text embeddings \(\mathbf{t}_j = f_T(s_j)\) for each class label \(s_j\). The similarity between image features and class text embeddings is computed, typically via cosine similarity. Applying a softmax normalization over classes yields a probability distribution over categories for each image:
\begin{equation}
    \hat{P}_{ij} = \frac{\exp(\tau \cdot \cos(\mathbf{v}_i, \mathbf{t}_j))}{\sum_{k=1}^{K} \exp(\tau \cdot \cos(\mathbf{v}_i, \mathbf{t}_k))},
\end{equation}
where \(\tau\) is a temperature parameter controlling the sharpness of the distribution. Stacking all \(P_{ij}\) forms the matrix \(\hat{\mathbf{P}} \in \mathbb{R}^{N \times K}\), representing the VLM-induced probability distributions over \(K\) classes for \(N\) images. Further, assuming that we have $V_2$ different VLMs, we obtain $V_2$ probability distributions
$\{\hat{\mathbf{P}}_v\}_{v=1}^{V_2}$. 

\subsubsection{Self-adaptive Optimal Transport}
Given $V_1$ probability distributions $\{\mathbf{P}_v\}_{v=1}^{V_1}$ obtained from VFMs and $V_2$ probability distributions $\{\hat{\mathbf{P}}_v\}_{v=1}^{V_2}$ obtained from VLMs, a straightforward integration strategy is to solve the following weighted OT problem:
\begin{equation}
\min_{\mathbf{T} \in \Pi(\mathbf{p}, \mathbf{q})} 
\sum_{v=1}^{V_1} \lambda_v \langle \mathbf{T}, \mathbf{C}_v \rangle
+ \sum_{v=1}^{V_2} \mu_v \langle \mathbf{T}, \hat{\mathbf{C}}_v \rangle
- \epsilon \mathcal{H}(\mathbf{T}),
\label{eq:weighted_ot_min}
\end{equation}
where $\mathbf{C}_v = \mathbf{E} - \mathbf{P}_v$ denote the cost matrices. $\lambda_v$~($\mu_v$) denote the weights assigned to the $v$-th VFM~(VLM). By substituting $\mathbf{T} \mathbf{1}  = \mathbf{p}$, problem~\eqref{eq:weighted_ot_min} can be equivalently reformulated as:
\begin{equation}
\max_{\mathbf{T} \in \Pi(\mathbf{p}, \mathbf{q})} 
\sum_{v=1}^{V_1} \lambda_v \langle \mathbf{T}, \mathbf{P}_v \rangle
+ \sum_{v=1}^{V_2} \mu_v \langle \mathbf{T}, \hat{\mathbf{P}}_v \rangle
+ \epsilon \mathcal{H}(\mathbf{T}).
\label{eq:weighted_ot_max}
\end{equation} 
Ideally, models with higher prediction quality should be assigned larger weights. However, in the zero-shot classification setting, ground-truth labels are unavailable, making it infeasible to determine these weights using a held-out validation set.

To overcome this limitation, we introduce \textbf{SOTA} (\textbf{S}elf-adaptive \textbf{O}ptimal \textbf{T}r\textbf{A}nsport), which automatically adjusts the contribution of each model without relying on labeled data. Specifically, we replace the Frobenius inner product \(\langle \cdot, \cdot \rangle\) with its squared form:
\begin{equation}
\max_{\mathbf{T} \in \Pi(\mathbf{p}, \mathbf{q})} \sum_{v=1}^{V_1} \langle \mathbf{T}, \mathbf{P}_v \rangle^2+\sum_{v=1}^{V_2} \langle \mathbf{T}, \hat{\mathbf{P}}_v \rangle^2 + \epsilon \mathcal{H}(\mathbf{T}).
\label{eq:selfadaptive_ot}
\end{equation}
In the next subsection, we show that the refined model can adaptively assign weights based on the transport distances between the plan \(\mathbf{T}\) and the set of cost matrices, thereby further improving the robustness of \textbf{SOTA} to less reliable or noisy distributions~({More discussions can be found in \emph{Appendix 6.2}}).

\paragraph{Extension.}  
While Eq.~\eqref{eq:selfadaptive_ot} provides a flexible way to ensemble multiple probability distributions via a unified transport plan, it has a key limitation: the visual distributions $\{\mathbf{P}_v\}$ from VFMs are learned independently of the semantic information in $\hat{\mathbf{P}}_v$ from VLMs. This decoupling often leads to poor alignment, as the GMM fitting for  $\{\mathbf{P}_v\}$ lacks semantic guidance. To address this, we propose a \textit{joint optimization} framework that simultaneously learns the GMM parameters \(\Theta=\{\mathbf\mu_c,\mathbf\Sigma_c,\pi_c\}\) and the transport plan $\mathbf{T}$:
\begin{equation}
\max_{\substack{\mathbf{T}\in\Pi(\mathbf{p},\mathbf{q}),\Theta}}
\sum_{v=1}^{V_1}\langle \mathbf{T}, \mathbf{P}_v(\Theta)\rangle^2
+\sum_{v=1}^{V_2}\langle \mathbf{T}, \hat{\mathbf{P}}_v\rangle^2
+\epsilon\,\mathcal{H}(\mathbf{T}),
\label{eq:selfadaptive_ot2}
\end{equation}
where \(\mathbf{P}_v(\Theta)\) denotes the VFM-induced posterior matrix that depends on the current GMM parameters \(\Theta\).
 In this coupled process, $\mathbf{T}$ is jointly shaped by the GMM-induced visual assignments and the VLM-derived semantic distributions, while also guiding the update of GMM parameters \(\Theta\). This mutual influence encourages the emergence of clusters that are both visually coherent and semantically consistent, leading to better alignment and more robust fusion.

\subsection{Optimization}

If the squared terms were absent (i.e., the objective were linear in \(\langle\mathbf{T},\cdot\rangle\)), the problem Eq.~\eqref{eq:selfadaptive_ot2} can be solved directly by an alternating optimization strategy. Specifically, we alternate between updating the assignment \(\langle\mathbf{T},\cdot\rangle\) ( via Sinkhorn algorithm~\cite{OT}) and updating the GMM parameters \(\Theta=\{\mathbf\mu_c,\mathbf\Sigma_c,\pi_c\}\) using the current assignments. However, the quadratic terms \(\langle\mathbf{T},\cdot\rangle^2\) nonlinearly couple \(\mathbf{T}\), so we resort to an iterative Minorization-Maximization (MM)~\cite{MM} scheme. The details can be found in \emph{Appendix 6.1}.

\begin{table*}[h]
\caption{Comparison with several state-of-the-art methods on 11 natural datasets under the \emph{Transductive} setting. For our method (\textbf{SOTA}), we employ different combinations of foundation models. Specifically, ‘CLIP-1’ and ‘CLIP-2’ denote the use of different visual encoders within CLIP (ViT-B/16 and RN50, respectively).}
\centering
\scriptsize
\renewcommand{\arraystretch}{1.3}
\setlength{\tabcolsep}{3.1pt}
\begin{tabular}{ l >{\columncolor{avgbg}}c *{11}{c} }
\toprule
 Method & \textbf{Average} & ImageNet & SUN397 & Aircraft & EuroSAT & StanfordCars & Food101 & Pets & Flower102 & Caltech101 & DTD & UCF101 \\
\hline
  CLIP-1~\cite{radford2021learning} & $65.2$ & 66.6 & 62.5 & 24.7 & 48.3 & 65.6 & 85.9 & 89.1 & 70.7 & 93.2 & 43.5 & 67.5 \\
  CLIP-2~\cite{radford2021learning} & $56.1$ & 58.2 & 58.8 & 15.7 & 23.7 & 55.7 & 74.0 & 83.6 & 61.8 & 85.9 & 40.4 & 58.8 \\
\hline
    GDA~\cite{GDA} & $67.6$\plus{2.4} & 67.3\plus{0.7} & 63.9\plus{1.4} & 25.5\plus{0.8} & 59.5\plus{11.2} & 67.0\plus{1.4} & 86.4\plus{0.5} & 90.8\plus{1.7} & 74.0\plus{3.3} & 93.8\plus{0.6} & 45.3\plus{1.8} & 70.3\plus{2.8}  \\
  $\text{ZLaP}$~\cite{kalantidis2024label}      & $67.5$\plus{2.3} & 69.7\plus{3.1} & 67.8\plus{5.3} & 26.3\plus{1.6} & 57.7\plus{9.4} & 66.8\plus{1.2} & 87.2\plus{1.3} & 87.9\minusv{-1.2} & 67.9\minusv{-2.8} & 91.8\minusv{-1.4} & 46.0\plus{2.5} & 73.8\plus{6.3} \\
  Stat.A~\cite{statA}    & $69.9$\plus{4.7} & 69.9\plus{3.3} & 68.7\plus{6.2} & 24.7\minusv{0.0} & 67.3\plus{19.0} & 68.0\plus{2.4} & 87.1\plus{1.2} & 92.4\plus{3.3} & 75.2\plus{4.5} & 94.2\plus{1.0} & 48.4\plus{4.9} & 73.5\plus{6.0} \\
  TransCLIP-2~\cite{transductive-4}  & $70.3$\plus{5.1} & 70.4\plus{3.8} & 68.9\plus{6.4} & 26.9\plus{2.2} & 66.1\plus{17.8} & 69.5\plus{3.9} & 87.1\plus{1.2} & 92.5\plus{3.4} & 76.5\plus{5.8} & 92.7\minusv{0.5} & 48.6\plus{5.1} & 74.1\plus{6.6} \\
  $\text{ECALP}$~\cite{li2025efficient} & $70.5$\plus{5.3} & 71.3\plus{4.7} & 70.4\plus{7.9} & 29.5\plus{4.8} & 56.5\plus{8.2} & 68.2\plus{2.6} & 85.7\minusv{-0.2} & 92.3\plus{3.2} & 76.0\plus{5.3} & 94.4\plus{1.2} & 56.3\plus{12.8} & 75.4\plus{7.9} \\
  $\text{ADAPT}$~\cite{zhang2025backpropagation} & $72.4$\plus{7.2} & 71.6\plus{5.0} & 72.3\plus{9.8} & 30.8\plus{6.1} & 65.9\plus{17.6} & 71.3\plus{5.1} & 85.1\minusv{-0.8} & 92.6\plus{3.5} & 80.1\plus{9.3} & 95.5\plus{2.3} & 56.9\plus{1.4} & 73.9\plus{6.4} \\
  GTA-CLIP~\cite{saha2025generate} & $74.5$\plus{9.3} & 71.9\plus{5.3} & 73.5\plus{11.0} & 29.3\plus{4.6} & 76.4\plus{28.1} & 72.1\plus{6.5} & 87.4\plus{1.5} & 93.4\plus{4.3} & 82.1\plus{11.4} & 95.5\plus{2.3} & 58.5\plus{15.0} & 79.1\plus{11.6} \\

\hline
  \textbf{SOTA} $\downarrow$ \\
  CLIP-1+CLIP-2 & $72.5$\plus{7.3} & 69.9\plus{3.3} & 70.2\plus{7.7} & 22.6\minusv{-2.1} & 82.9\plus{34.6} & 69.9\plus{4.3} & 86.0\plus{0.1} & 92.4\plus{3.3} & 75.1\plus{4.4} & 92.9\minusv{-0.3} & 56.0\plus{12.5} & 79.4\plus{11.9} \\
  CLIP-1+DINOv2 & $75.7$\plus{10.5} & 77.7\plus{11.1} & 71.9\plus{9.4} & 28.0\plus{3.3} & 84.4\plus{36.1} & 73.9\plus{8.3} & 89.0\plus{3.1} & 94.5\plus{5.4} & 83.9\plus{13.2} & 96.1\plus{2.9} & 52.8\plus{9.3} & 80.3\plus{12.8} \\
  CLIP-1+DINOv3 & $\underline{77.4}$\plus{12.2} & 78.7\plus{12.1} & 72.6\plus{10.1} & 33.0\plus{8.3} & 86.4\plus{38.1} & 79.6\plus{14.0} & 90.1\plus{4.2} & 95.1\plus{6.0} & 84.2\plus{13.5} & 95.9\plus{2.7} & 56.8\plus{13.3} & 79.2\plus{11.7} \\
  CLIP-1+DINO(v2+v3) & $\textbf{77.8}$\plus{12.6} & 79.9\plus{13.3} & 72.8\plus{10.3} & 34.0\plus{9.3} & 87.2\plus{38.9} & 78.5\plus{12.9} & 90.0\plus{4.1} & 95.3\plus{6.2} & 84.5\plus{13.8} & 96.6\plus{3.4} & 55.9\plus{12.4} & 81.1\plus{13.6} \\
\bottomrule
\end{tabular}
\label{tab:trans_nature}
\end{table*}

\begin{table*}[h]
\caption{Comparison with TransCLIP~\cite{transductive-4}, one of the leading competitors, on 10 remote sensing datasets under the \emph{Transductive} setting. Methods with the prefix “T-” indicate applying TransCLIP~\cite{transductive-4} to the original foundation models.}
\centering
\scriptsize
\renewcommand{\arraystretch}{1.3}
\setlength{\tabcolsep}{3.8pt}
\begin{tabular}{ l >{\columncolor{avgbg}}c *{10}{c} }
\toprule
 Method & \textbf{Average} & AID & EuroSAT & MLRSNet & OPTIMAL31 & PatternNet & RESISC45 & RSC11 & RSICB128 & RSICB256 & WHURS19 \\
\hline
    CLIP~\cite{radford2021learning}& $56.1$ & 66.4 & 45.3 & 51.2 & 73.0 & 59.6 & 60.7 & 55.5 & 27.7 & 40.3 & 81.1   \\
    GeoRSCLIP~\cite{georsclip}        & $64.5$ & 70.3 & 53.4 & 65.0 & 79.6 & 75.8 & 68.8 & 68.3 & 29.0 & 46.5 & 88.8 \\
    RemoteCLIP~\cite{remoteclip}       & $61.0$ & 91.7 & 35.5 & 56.3 & 77.6 & 55.9 & 68.1 & 61.8 & 26.0 & 41.5 & 95.2 \\
    SkyCLIP50~\cite{skyscript}        & $64.4$ & 70.3 & 52.6 & 63.2 & 79.5 & 73.8 & 66.7 & 61.2 & 39.0 & 47.1 & 91.0 \\
\hline
    T-CLIP          & $66.4$\plus{10.3} & 80.7\plus{14.3} & 49.0\plus{3.7} & 64.2\plus{13.0} & 82.9\plus{9.9} & 76.6\plus{17.0} & $74.1$\plus{13.4} & 67.0\plus{11.5} & 33.2\plus{5.5} & 46.4\plus{6.1} & 90.3\plus{9.2} \\
    T-GeoRSCLIP     & $\underline{76.2}$\plus{11.7} & 78.2\plus{7.9} & 69.0\plus{15.6} & 71.9\plus{6.9} & 87.3\plus{7.7} & 94.5\plus{18.7} & $79.5$\plus{10.7} & 78.6\plus{10.3} & 42.8\plus{13.8} & 61.8\plus{15.3} & 98.7\plus{9.9} \\
    T-RemoteCLIP    & $70.8$\plus{9.8} & 95.6\plus{3.9} & 51.0\plus{15.5} & 65.8\plus{9.5} & 87.8\plus{10.2} & 70.7\plus{14.8} & $79.4$\plus{11.3} & 79.7\plus{17.9} & 31.1\plus{5.1} & 49.2\plus{7.7} & 97.9\plus{2.7} \\
    T-SkyCLIP50     & $75.0$\plus{10.6} & 78.7\plus{8.4} & 64.5\plus{11.9} & 73.2\plus{10.0} & 85.2\plus{5.7} & 87.6\plus{13.8} & $77.3$\plus{10.6} & 77.1\plus{15.9} & 49.4\plus{10.4} & 59.1\plus{12.0} & 97.8\plus{6.8} \\
\hline
    SOTA & $\textbf{81.5}$\plus{17.0} & 96.6\plus{4.9} & 71.7\plus{18.3} & 81.1\plus{16.1} & 93.0\plus{13.4} & 96.5\plus{20.7} & 88.8\plus{20.0} & 78.8\plus{10.5} & 47.6\plus{8.6} & 61.1\plus{14.0} & 99.3\plus{4.1} \\
\bottomrule
\end{tabular}
 \vspace{-0.5cm}
\label{tab:trans_sensing}
\end{table*}


\section{Experiments}
To evaluate the effectiveness of our method, we conducted experiments on three distinct types of datasets: natural images, remote sensing data, and medical imaging data. More details can be found in \emph{Appendix 7}.

\subsection{Main Results}
We evaluate under two settings: \emph{Transductive inference} and \emph{Inductive inference}. In the \emph{Transductive inference} setting, $\mathcal{D}_u$ itself constitutes the entire test set. In the \emph{Inductive inference} setting, $\mathcal{D}_u$ is regarded as training data. 
The goal is to learn a classifier that generalizes to unseen test instances from the same distribution. 

\paragraph{Transductive setting.}
Table \ref{tab:trans_nature}, Table \ref{tab:trans_sensing}, and Table \ref{tab:trans_medicine} report main results on datasets from different domains. It should be noted that for remote sensing and medical data, we only compare with TransCLIP~\cite{transductive-4}, as it has been demonstrated to be one of the leading competitors on natural image datasets. The experimental results are drawn from~\cite{transclip-GS,transclip-m}. Observing the comparative results, we can draw two important conclusions:
\begin{itemize}
    \item \textbf{Effectiveness of transductive methods.}
As shown in Table~\ref{tab:trans_nature}, all transductive approaches significantly improve prediction accuracy compared with the original CLIP. 
This improvement mainly stems from the effective utilization of visual feature distributions during the label correction process. 
To some extent, the performance gain is positively correlated with the quality of the visual representations.  For instance, when a stronger visual encoder such as DINOv3 is adopted, our method (\textbf{SOTA}) achieves superior results, surpassing the best competitor by an average margin of approximately 6.9\%, which demonstrates its ability to fully exploit high-quality visual features.
    \item \textbf{Complementarity across different VLMs.}
The performance of different VLMs varies considerably across datasets due to discrepancies in their pre-training strategies. 
On natural image datasets, CLIP-1 outperforms CLIP-2 mainly because the ViT-B/16 encoder produces more discriminative visual features than RN50. 
By integrating the complementary strengths of multiple foundation models, our approach consistently outperforms individual models in most cases, 
highlighting the effectiveness of leveraging multiple VLMs for zero-shot generalization.
\end{itemize}

\begin{table}[h]
\caption{Comparison with TransCLIP~\cite{transductive-4}, one of the leading competitors, on 5 medical datasets under the \emph{Transductive} setting. Methods with the prefix “T-” indicate applying TransCLIP~\cite{transductive-4} to the original foundation models.}
\centering
\scriptsize
\renewcommand{\arraystretch}{1.3}
\setlength{\tabcolsep}{2pt}
\begin{tabular}{c >{\columncolor{avgbg}}c *{5}{c}}
\toprule
\multicolumn{1}{c}{Method} & 
\multicolumn{1}{c}{\textbf{Average}} & 
\raisebox{-.5\height}{\rotatebox{50}{SICAP-MIL}} & 
\raisebox{-.5\height}{\rotatebox{50}{SKINCANCER}} & 
\raisebox{-.5\height}{\rotatebox{50}{LC-LUNG}} & 
\raisebox{-.5\height}{\rotatebox{50}{NCT-CRC}} & 
\raisebox{-.5\height}{\rotatebox{50}{WSSS4LUAD}} \\
\hline
CLIP~\cite{radford2021learning} & $30.8$ & 29.8 & 3.6 & 31.3 & 24.4 & 64.9  \\
CONCH~\cite{conchlu2024visual} & $62.9$ & 27.4 & 53.7 & 90.0 & 61.3 & 82.2 \\
PLIP~\cite{pliphuang2023visual} & $58.2$ & 47.3 & 22.4 & 85.0 & 63.0 & 73.1 \\
MUSK~\cite{muskxiang2025vision} & $66.3$ & 36.6 & 59.6 & 92.9 & 66.9 & 75.4 \\
\hline
T-CLIP & $34.7$\plus{3.9} & 25.2\minusv{-4.6} & 11.4\plus{7.8} & 29.6\minusv{-1.7} & 40.4\plus{16.0} & 66.7\plus{1.8} \\
T-CONCH & $68.1$\plus{5.2} & 29.4\plus{2.0} & 63.7\plus{10.0} & 96.3\plus{6.3} & 66.4\plus{5.1} & 84.9\plus{2.7} \\
T-PLIP & $69.4$\plus{11.2} & 53.3\plus{6.0} & 37.0\plus{14.6} & 93.6\plus{8.6} & 78.1\plus{15.1} & 85.0\plus{11.9} \\
T-MUSK & $\underline{76.0}$\plus{9.7} & 39.4\plus{2.8} & 64.8\plus{5.2} & 97.2\plus{4.3} & 90.6\plus{23.7} & 88.2\plus{12.8} \\
\hline
\textbf{SOTA} & $\textbf{83.9}$\plus{14.1} & 65.5\plus{18.2} & 78.9\plus{19.3} & 97.7\plus{4.8} & 86.9\plus{20.0} & 90.4\plus{8.2}  \\
\bottomrule
\end{tabular}
 \vspace{-0.5cm}
\label{tab:trans_medicine}
\end{table}

\begin{table*}[h]
\caption{Comparison with several state-of-the-art methods on 11 natural datasets under \textbf{Inductive} settings. * indicates TTA method.}
\centering
\scriptsize
\renewcommand{\arraystretch}{1.3}
\setlength{\tabcolsep}{3pt}
\begin{tabular}{ l >{\columncolor{avgbg}}c *{11}{c} }
\toprule
 Method & \textbf{Average} & ImageNet & SUN397 & Aircraft & EuroSAT & StanfordCars & Food101 & Pets & Flower102 & Caltech101 & DTD & UCF101 \\
\hline
  CLIP-1~\cite{radford2021learning} & $65.2$ & 66.6 & 62.5 & 24.7 & 48.3 & 65.6 & 85.9 & 89.1 & 70.7 & 93.2 & 43.5 & 67.5 \\
  CLIP-2~\cite{radford2021learning} & $56.1$ & 58.2 & 58.8 & 15.7 & 23.7 & 55.7 & 74.0 & 83.6 & 61.8 & 85.9 & 40.4 & 58.8 \\
\hline

    Zero~\cite{ZERO} & $66.0$\plus{0.8} & 71.2\plus{4.6} & 67.6\plus{5.1} & 25.2\plus{0.5} & 42.2\minusv{-6.1} & 69.0\plus{3.4} & 86.8\plus{0.9} & 87.8\minusv{-1.3} & 67.2\minusv{-3.5} & 94.4\plus{1.2} & 45.9\plus{2.4} & 69.2\plus{1.7} \\
    MTA*~\cite{MTA} & $66.0$\plus{0.8} & 70.1\plus{3.5} & 66.7\plus{4.2} & 25.2\plus{0.5} & 45.4\minusv{-2.9} & 68.5\plus{2.9} & 85.0\minusv{-0.9} & 88.2\minusv{-0.9} & 68.1\minusv{-2.6} & 94.2\plus{1.0} & 45.9\plus{2.4} & 68.7\plus{1.2} \\
    TDA*~\cite{karmanov2024efficient} & $67.7$\plus{2.5} & 69.5\plus{2.9} & 67.6\plus{5.1} & 23.9\minusv{-0.8} & 58.0\plus{9.7} & 67.3\plus{1.7} & 86.1\plus{0.2} & 88.6\minusv{-0.5} & 71.4\plus{0.7} & 94.2\plus{1.0} & 47.4\plus{3.9} & 70.7\plus{3.2} \\
    TIPPLE*~\cite{TIPPLE} & $67.8$\plus{2.6} & 71.0\plus{4.4} & 68.1\plus{5.6} & 25.4\plus{0.7} & 51.8\plus{3.5} & 67.8\plus{2.2} & 86.0\plus{0.1} & 90.2\plus{1.1} & 71.3\plus{0.6} & 93.9\plus{0.7} & 49.2\plus{5.7} & 71.2\plus{3.7} \\
    DPE*~\cite{DPE} & $69.6$\plus{4.4} & 71.9\plus{5.3} & 70.1\plus{7.6} & 29.0\plus{4.3} & 55.8\plus{7.5} & 67.3\plus{1.7} & 86.2\plus{0.3} & 91.1\plus{2.0} & 75.1\plus{4.4} & 94.8\plus{1.6} & 54.2\plus{10.7} & 70.4\plus{2.9} \\
    DMN*~\cite{zhang2024dual} & $70.5$\plus{5.3} & 72.2\plus{5.6} & 70.2\plus{7.7} & 30.0\plus{5.3} & 59.4\plus{11.1} & 68.0\plus{2.4} & 85.1\minusv{-0.8} & 92.0\plus{2.9} & 74.5\plus{3.8} & 95.4\plus{2.2} & 55.8\plus{12.3} & 72.5\plus{5.0} \\
    COSMIC*~\cite{huang2025cosmic} & $73.3$\plus{8.1} & 78.2\plus{11.6} & 72.3\plus{9.8} & 31.4\plus{6.7} & 58.8\plus{10.5} & 71.3\plus{5.7} & 86.6\plus{0.7} & 94.2\plus{5.1} & 82.1\plus{11.4} & 96.8\plus{3.6} & 58.2\plus{14.7} & 76.2\plus{8.7} \\
\hline
\textbf{SOTA} $\downarrow$ \\
  CLIP-1+CLIP-2 & $71.5$\plus{6.3} & 68.9\plus{2.3} & 69.8\plus{7.3} & 21.8\minusv{-2.9} & 82.5\plus{34.2} & 68.5\plus{2.9} & 86.0\plus{0.1} & 91.6\plus{2.5} & 76.7\plus{6.0} & 93.3\plus{0.1} & 52.1\plus{8.6} & 75.8\plus{8.3}  \\
  CLIP-1+DINOv2 & $75.3$\plus{10.1} & 77.0\plus{10.4} & 71.9\plus{9.4} & 25.4\plus{0.7} & 84.3\plus{36.0} & 73.8\plus{8.2} & 89.0\plus{3.1} & 94.4\plus{5.3} & 83.6\plus{12.9} & 96.6\plus{3.4} & 53.7\plus{10.2} & 79.1\plus{11.6} \\
  CLIP-1+DINOv3 & $\underline{77.2}$\plus{12.0} & 77.9\plus{11.3} & 72.3\plus{9.8} & 31.8\plus{7.1} & 85.9\plus{37.6} & 78.8\plus{13.2} & 90.0\plus{4.1} & 95.1\plus{6.0} & 84.1\plus{13.4} & 95.2\plus{2.0} & 57.5\plus{14.0} & 80.3\plus{12.8} \\
  CLIP-1+DINO(v2+v3) & $\textbf{77.6}$\plus{12.4} & 79.1\plus{12.5} & 73.0\plus{10.5} & 31.5\plus{6.8} & 87.2\plus{38.9} & 77.5\plus{11.9} & 90.2\plus{4.3} & 95.4\plus{6.3} & 85.1\plus{14.4} & 96.9\plus{3.7} & 56.4\plus{12.9} & 81.7\plus{14.2} \\
\bottomrule
\end{tabular}
 \vspace{-0.5cm}
\label{tab:induc_nature}
\end{table*}

\paragraph{Inductive setting.}
Under the \emph{Inductive} setting, we compare our proposed method with several representative test-time adaptation approaches, 
which either perform re-learning based on test data or apply label correction using pre-stored samples during testing. 
As shown in Table~\ref{tab:induc_nature}, our method (\textbf{SOTA}) achieves comparable or even superior performance on the majority of datasets 
compared with existing test-time adaptation methods. This demonstrates that the ensemble classifier learned on the training data exhibits strong generalization capability to unseen data. Both DMN and COSMIC achieve results comparable to ours; however, they require maintaining a cached sequence during testing, leading to higher storage consumption. In contrast, our method achieves sample-wise inference at test time with only a few GMM parameters and adaptive coefficients. 


\subsection{Ablation studies}
\begin{figure}[h]
    \centering
    \includegraphics[width=0.47\textwidth]{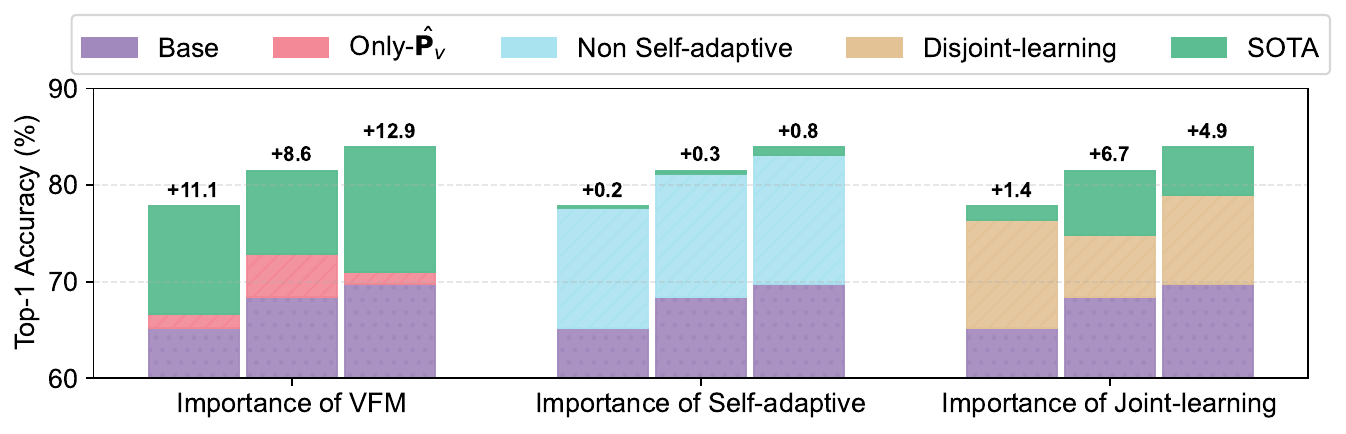}
    \caption{Ablation experiments showing top-1 accuracy (\%) across three dataset types: each group consists of three bars (from left to right: natural, remote sensing, and medical images). The bars correspond to the following settings: \textbf{Base}, \textbf{Only-$\hat{\mathbf{P}}_v$}, \textbf{Non Self-adaptive}, \textbf{Disjoint-learning}, and \textbf{SOTA}. }
     \vspace{-0.5cm}
    \label{Fig:ablation}
\end{figure}

We conduct a comprehensive ablation study to verify the effectiveness of each component in our framework, as shown in Fig.~\ref{Fig:ablation}. Specifically, \textbf{Base} denotes the performance of the baseline model. It should be noted that for remote sensing and medical datasets, this represents the highest performance achieved among base models. \textbf{Only-$\hat{\mathbf{P}}_v$} performs Optimal Transport solely based on the semantic probability distribution $\hat{\mathbf{P}}_v$ output by the VLMs, without utilizing any VFMs. \textbf{Non Self-adaptive} indicates that both $\lambda$ and $\mu$ remain fixed at their initial values throughout the iterative process. \textbf{Disjoint-learning} performs inference by decoupling the optimization of GMM from the iterative process, with the aim of validating the importance of coupled learning.

\paragraph{How important is VFM introduction?}
The suboptimal performance of VLMs can be mainly attributed to two key factors. 
First, the training paradigm constrains the capability of visual encoders to extract high-quality image representations, as shown in Fig.~\ref{Fig:clip_v1_v2}. 
Second, there exists an inherent discrepancy between the visual and linguistic modalities. 
Consequently, it becomes crucial during inference to leverage the correlations between samples and categories derived from VFMs to rectify the predictions of VLMs, as such correlations tend to be more reliable and semantically grounded. 
As shown in Fig.~\ref{Fig:ablation}, while integrating the probabilistic outputs of multiple VLMs through our proposed SOTA ensemble strategy can yield improvements over a single model, the performance gain remains considerably smaller than that achieved when incorporating information from visual foundation models. Further analyses can be found in \emph{Appendix 8.1}.

\paragraph{How important is self-adaptive optimal transport?}
Considering the capability discrepancies among different foundation models, the SOTA ensemble employs a self-adaptive OT mechanism to automatically balance the cost matrices contributed by each model during the fusion process. 
As shown in Fig.~\ref{Fig:ablation}, the adaptive weighting scheme achieves better average performance across diverse datasets compared to the fixed-weight strategy.  We attribute the relatively modest improvement to two main factors: 
(1) The performance gaps among different foundation models, especially among visual foundation models, may be small on some datasets, and the differences mainly arise from modality-specific variations rather than intrinsic model diversity; 
and (2) when multiple strong foundation models are present, the influence of weaker models tends to be suppressed. 
To further substantiate this observation, we conduct experiments using CLIP~\cite{radford2021learning} and MUSK~\cite{muskxiang2025vision} as baseline models, chosen for their considerable performance disparity. As shown in Table~\ref{tab:adaptive}, both the Non-adaptive and Adaptive variants achieve substantial improvements over the baselines. However, the non-adaptive approach consistently lags behind the self-adaptive one across all datasets, underscoring the crucial role of the adaptive mechanism. Overall, these results confirm that the self-adaptive optimal transport strategy delivers a more pronounced performance gain compared to fixed-weighting schemes under this setting.
\begin{table}[h]
\caption{Ablation studies on self-adaptive optimal transport.}
\centering
\scriptsize
\renewcommand{\arraystretch}{1.3}
\setlength{\tabcolsep}{2pt}
\begin{tabular}{c >{\columncolor{avgbg}}c *{5}{c}}
\toprule
\multicolumn{1}{c}{Method} & 
\multicolumn{1}{c}{\textbf{Average}} & 
\raisebox{-.5\height}{\rotatebox{50}{SICAP-MIL}} & 
\raisebox{-.5\height}{\rotatebox{50}{SKINCANCER}} & 
\raisebox{-.5\height}{\rotatebox{50}{LC-LUNG}} & 
\raisebox{-.5\height}{\rotatebox{50}{NCT-CRC}} & 
\raisebox{-.5\height}{\rotatebox{50}{WSSS4LUAD}} \\
\hline
Non-adaptive & $71.8$ & 42.2 & 61.1 & 96.2 & 77.7 & 81.9  \\
Self-adaptive & $73.1$ & 45.9 & 62.5 & 96.3 & 78.1 & 82.5 \\
\bottomrule
\end{tabular}
 \vspace{-0.5cm}

\label{tab:adaptive}
\end{table}


\paragraph{How important is joint-learning?}
Our method adopts a joint optimization objective that couples the optimization of $\mathbf{T}$ with the visual assignments $\mathbf{P}_v(\Theta)$, enabling their co-evolution. The ablation results in Fig.~\ref{Fig:ablation} highlight the effectiveness of this coupled strategy: although the decoupled learning approach achieves a noticeable improvement over the base model, it still falls short of the coupled learning paradigm. 
This observation confirms that the joint-learning framework facilitates mutual reinforcement between the two components. 
Specifically, the probabilistic distribution estimated by the GMM from VFMs improves the accuracy of consistency-based predictions, 
while the latter, in turn, provides more reliable supervisory signals that guide the parameter estimation of the GMM.

\begin{figure}[h!]
    \centering
    \includegraphics[width=0.5\textwidth]{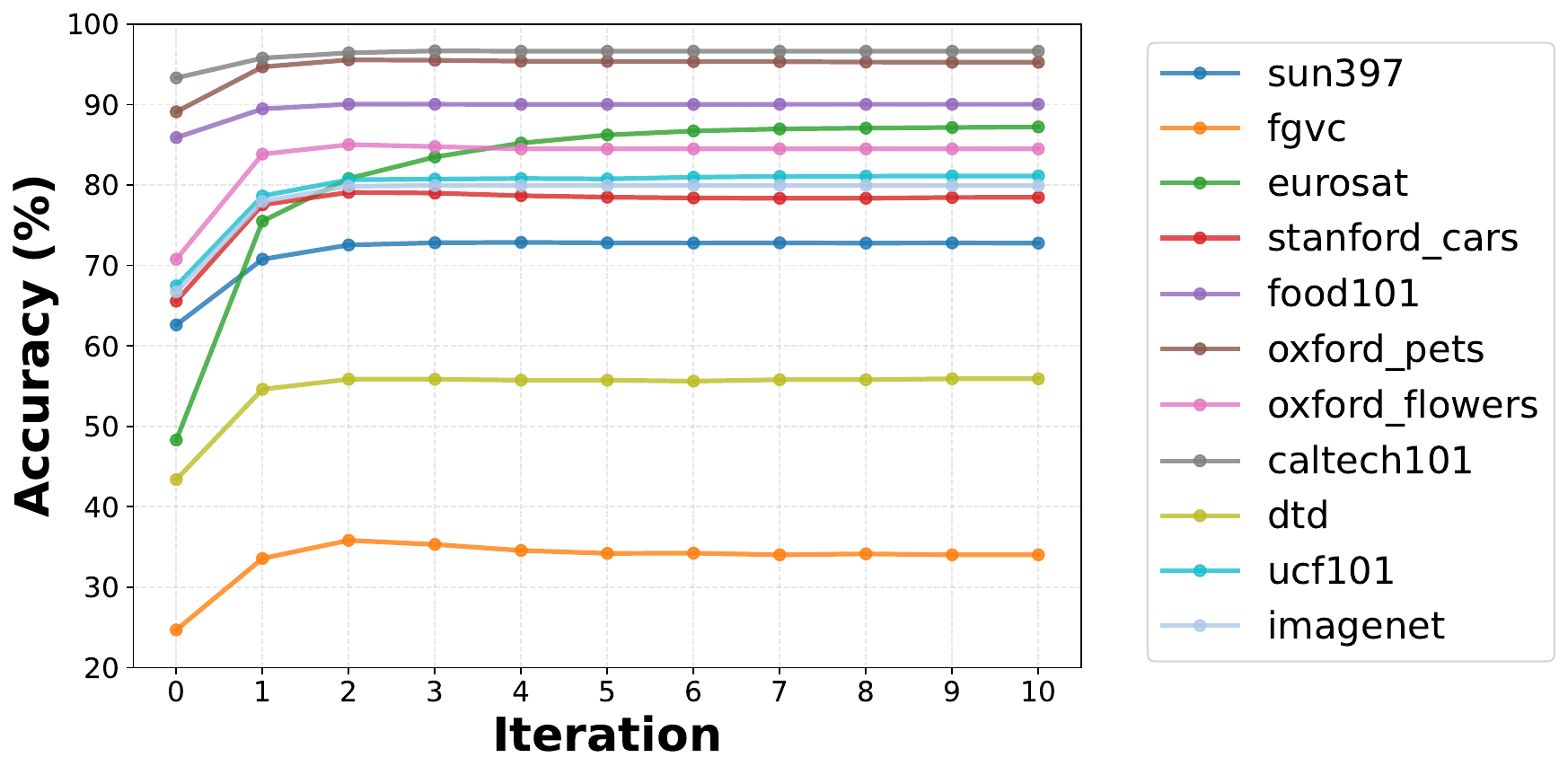}
    \caption{Convergence curves on 11 natural datasets. Our method often converges within a few iterations.}
    \vspace{-0.5cm}
    \label{Fig:train}
\end{figure}
\subsection{Convergence analysis}
We adopt an iterative MM algorithm to optimize Eq.~\eqref{eq:selfadaptive_ot2}. 
To assess its convergence properties, we evaluate the algorithm across eleven natural image datasets. 
As illustrated in Fig.~\ref{Fig:train}, the proposed method rapidly reaches a stable state after approximately five iterations on most datasets and achieves full convergence within ten iterations. 
Importantly, the iterative procedure of \textbf{SOTA} relies solely on basic matrix operations, which contributes to its low computational overhead. 
These observations highlight the efficiency and practicality of our approach, demonstrating that reliable results can be obtained with minimal computational overhead.

\section{Conclusion}
In this work, we presented \textbf{SOTA}, a \textit{training-free} ensemble framework that adaptively integrates the outputs of multiple foundation models through self-adaptive optimal transport. Without requiring any fine-tuning or hyperparameter adjustment, SOTA effectively balances model contributions and leverages the complementary strengths of different VFMs and VLMs. Extensive experiments across natural, medical, and remote sensing domains demonstrate its strong generalization and consistent performance gains over individual models. 

In future work, we plan to further enhance the ensemble mechanism along two directions. First, we aim to develop more sophisticated ensemble learning algorithms to better integrate the diverse predictions of VLMs and exploit their latent complementarity at the semantic level. 
Second, by treating each VFM as a distinct view, we will explore advanced multi-view clustering strategies to derive more reliable probabilistic representations, which can in turn provide stronger guidance for prediction correction.


\section*{Acknowledgments}
This work is supported by the Basic Research Project of Yunnan Province(Grant No. 202501CF070004),  Xingdian Talent Support Program, and Intelligent Computing Center, Yunnan Normal University.

\maketitlesupplementary

\section*{Contents}  
\addcontentsline{toc}{section}{Table of Contents for Supplementary Materials}  

\begin{itemize}[label={}]
    \item \textbf{6. Optimization} \hfill \pageref{sec:optimization}
        \begin{itemize}[label={}]
            \item \textbf{6.1 Optimization procedure} \hfill \pageref{sec:optimization_procedure}  
            \item \textbf{6.2 Discussion} \hfill
            \pageref{sec:optimization_discuss}  
        \end{itemize}  
    
    \item \textbf{7. Experimental Setup} \hfill \pageref{sec:setup}
        \begin{itemize}[label={}]
            \item \textbf{7.1 The details of datasets} \hfill \pageref{sec:datasets}  
            \item \textbf{7.2 Implementation details} \hfill
            \pageref{sec:details}  
        \end{itemize}  
    
    \item \textbf{8. Extended Discussions} \hfill \pageref{sec:discussion}
        \begin{itemize}[label={}]
            \item \textbf{8.1 How important is VFM introduction?} \hfill \pageref{sec:vfms}  
            \item \textbf{8.2 Qualitative Visualization and Analysis} \hfill \pageref{sec:visualization}
            \item \textbf{8.3 Computational efficiency analysis} \hfill \pageref{sec:efficiency}
            \item \textbf{8.4 Model subset analysis} \hfill \pageref{sec:subset}
        \end{itemize}  
\end{itemize}  

\section{Optimization}
\label{sec:optimization}

We propose a \textit{joint optimization} framework that simultaneously learns the GMM parameters \(\Theta=\{\mathbf\mu_c,\mathbf\Sigma_c,\pi_c\}\) and the transport plan $\mathbf{T}$:
\begin{equation}
\max_{\substack{\mathbf{T}\in\Pi(\mathbf{p},\mathbf{q}),\Theta}}
\sum_{v=1}^{V_1}\langle \mathbf{T}, \mathbf{P}_v(\Theta)\rangle^2
+\sum_{v=1}^{V_2}\langle \mathbf{T}, \hat{\mathbf{P}}_v\rangle^2
+\epsilon\,\mathcal{H}(\mathbf{T}),
\label{eq:original_objective}
\end{equation}

\subsection{Optimization procedure}
\label{sec:optimization_procedure}

\begin{theorem}
\label{thm:quadratic_minorizer}
Let $f(x) = x^2$ be a convex function and let $x^{(k)} \in \mathbb{R}$ be a given point.  
Then the first-order Taylor expansion of $f(x)$ at $x^{(k)}$ yields the global lower bound
\begin{equation}
  x^2 \ge (x^{(k)})^2 + 2x^{(k)}\left(x - x^{(k)}\right)
  = 2x^{(k)} x - (x^{(k)})^2,
\end{equation}
with equality if and only if $x = x^{(k)}$.  
Moreover, this affine function serves as a valid \emph{minorizer} of $f(x)$ that can be maximized in iterative optimization schemes.
\end{theorem}

Let \(\mathbf{P}_v^{(k)}=\mathbf{P}_v(\Theta^{(k)})\) be the posteriors computed from the current GMM parameters. We have

\begin{equation}
a_{P,v}^{(k)} \coloneqq \langle \mathbf{T}^{(k)}, \mathbf{P}_v^{(k)}\rangle,\qquad
a_{\hat{P},v}^{(k)} \coloneqq \langle \mathbf{T}^{(k)}, \hat{\mathbf{P}}_v\rangle,    
\end{equation}
According to Theorem~\ref{thm:quadratic_minorizer}, applying  tangent minorization to every quadratic term yields the surrogate (up to additive constants independent of \(\mathbf{T}\) and \(\Theta\)):
\begin{equation}\label{eq:surrogate}
\begin{split}
 G^{(k)}(\mathbf{T},\Theta)
\;=\;
&\sum_{v=1}^{V_1} 2 a_{P,v}^{(k)} \langle \mathbf{T}, \mathbf{P}_v(\Theta)\rangle\\
&+\sum_{v=1}^{V_2} 2 a_{\hat{P},v}^{(k)} \langle \mathbf{T}, \hat{\mathbf{P}}_v\rangle
+\epsilon\,\mathcal{H}(\mathbf{T}).   
\end{split}
\end{equation}
Hence, at iteration \(k\) we maximize the surrogate \(G^{(k)}\), which is linear in \(\mathbf{T}\) for fixed \(\Theta\). In practice we optimize \(G^{(k)}\) by alternating updates:

\begin{itemize}
  \item \textbf{(Update \(\mathbf{T}\) given \(\Theta^{(k)}\)).} 
    With \(\mathbf{P}_v=\mathbf{P}_v(\Theta^{(k)})\) fixed, define adaptive weights
    \begin{equation}\label{eq:weight}
       \lambda_{v}^{(k)} \coloneqq 2 a_{P,v}^{(k)},\qquad
    \mu_{v}^{(k)} \coloneqq 2 a_{\hat{P},v}^{(k)}.  
    \end{equation}
   
    The \(\mathbf{T}\)-subproblem becomes
    \begin{equation}
         \max_{\mathbf{T}\in\Pi(\mathbf{p},\mathbf{q})}
    \Big\langle \mathbf{T}, \; \mathbf{S}^{(k)} \Big\rangle
    + \epsilon\,\mathcal{H}(\mathbf{T}),  
    \end{equation}
where 
\begin{equation}
     \mathbf{S}^{(k)} \coloneqq \sum_{v=1}^{V_1}\lambda_{v}^{(k)}\mathbf{P}_v
    +\sum_{v=1}^{V_2}\mu_{v}^{(k)}\hat{\mathbf{P}}_v.   
\label{eq:14}
\end{equation}

This is a classical entropic-regularized optimal transport problem, which can be efficiently solved using the Sinkhorn algorithm~\cite{OT}. Concretely, form the kernel \(\mathbf{K}=\exp(\mathbf{S}^{(k)}/\epsilon)\) and compute
\begin{equation}
\label{eq:sinkhorn}
\mathbf{T}^{(k+1)} = \mathrm{Diag}(\mathbf{p}) \cdot \exp\left( \frac{\mathbf{S}^{(k)}}{\epsilon} \right) \cdot \mathrm{Diag}(\mathbf{q}),
\end{equation}
where $\mathbf{p} \in \mathbb{R}^K$ and $\mathbf{q} \in \mathbb{R}^N$ are scaled to satisfy the marginals by the usual Sinkhorn updates
\begin{align}
\mathbf{p}^{(s+1)} &= \frac{\mathbf{1}_K}{\mathbf{K} \mathbf{q}^{(s)}}, \\
\mathbf{q}^{(s+1)} &= \frac{\mathbf{1}_N}{\mathbf{K}^\top \mathbf{p}^{(s+1)}},
\end{align}
It is worth noting that once $\mathbf{T}$ is updated, we re-update the weight parameters according to Eq.~\eqref{eq:weight}.

  \item \textbf{(Update GMM parameters \(\Theta\) given \(\mathbf{T}^{(k+1)}\)).}
    We treat \(\mathbf{T}^{(k+1)}\) as soft assignments of samples to classes and perform an M-step update for the GMM parameters. Specifically,

\begin{equation}\label{eq:gmm1}
\mathbf{\mu}_c^{(k+1)} = \frac{\sum_i T_{ic}^{(k+1)} \mathbf{v}_i}{\sum_i Q_{ic}^{(t)}},
\end{equation}
\begin{equation}\label{eq:gmm2}
\mathbf{\Sigma}^{(k+1)} = \frac{1}{N} \sum_i \sum_c T_{ic}^{(k+1)} (\mathbf{v}_i - \mathbf{\mu}_c^{(k+1)})(\mathbf{v}_i - \mathbf{\mu}_c^{(k+1)})^\top,
\end{equation}
where we omit the GMM index for brevity, though these updates are independently applied to each visual model’s GMM. After the M-step we recompute the posterior matrices \(\mathbf{P}_v(\Theta^{(k+1)})\) (E-step).
\end{itemize}

The full iterative procedure alternates the two blocks above~(See Algorithm~\ref{alg:sota}). 
Because each \(\mathbf{T}\)-update maximizes the surrogate \(G^{(k)}(\cdot,\Theta^{(k)})\) and the construction of \(G^{(k)}\) is a valid minorizer of the original objective~\ref{eq:original_objective}, the algorithm yields a non-decreasing sequence of objective values and converges to a stationary point.

\begin{algorithm}[h]
	\caption{Optimization of \textbf{SOTA} via MM}
	\label{alg:sota}
	\begin{algorithmic}[1]
		\Require Cost matrices $\{\mathbf{C}_v\}_{v=1}^{V_1+V_2}$, marginal distributions $\mathbf{p}, \mathbf{q}$, visual features $\{\mathbf{v}_v\}_{v=1}^{V_1+V_2}$, entropy weight $\epsilon$, and the number of maximum iteration $T$.
		\Ensure Optimal transport plan $\mathbf{T}^*$
		\State Initialize transport plan $\mathbf{T}^{(0)}$.
		\State Initialize $\Theta^{(0)}, \lambda^{(0)}_v, \mu^{(0)}_v$.
		\For{$k = 1$ to $T$}
		\State \textcolor{gray}{\# Step 1: Compute posterior probability.}
		\For{$v = 1$ to $V_1 + V_2$}
		\State $\mathbf{P}_v^{(k)}(\Theta) \leftarrow  (\Theta^{(k)}, \mathbf{v}_v) $
		\Comment{GMM E-step}
		\EndFor
		
		
		\State \textcolor{gray}{\# Step 2: Update transport plan.}
		
		\State $\mathbf{T}^{(k+1)} = \mathrm{Diag}(\mathbf{p}) \cdot \exp\left( \frac{\mathbf{S}^{(k)}}{\epsilon} \right) \cdot \mathrm{Diag}(\mathbf{q})$ \\
		\Comment{Eq.(\ref{eq:sinkhorn})}
		
		\State \textcolor{gray}{\# Step 3: Update adaptive weights.}
		\State $\lambda^{(k+1)}_v \leftarrow (\mathbf{T}^{(k+1)}, \mathbf{P}^{(k)}_v)$
		\State $\mu^{(k+1)}_v \leftarrow (\mathbf{T}^{(k+1)}, \mathbf{\hat{P}}^{(k)}_v)$
		\Comment{Eq.(\ref{eq:weight})}

		\State \textcolor{gray}{\# Step 4: Update GMM.}
		\State $\Theta^{(k+1)} \leftarrow (\mathbf{T}^{(k+1)}, \{\mathbf{v}_v\}_{v=1}^{V_1+V_2})$
		\Comment{Eq.(\ref{eq:gmm1}) and Eq.(\ref{eq:gmm2})}

		\EndFor
		\State \Return $\mathbf{T}^* \leftarrow \mathbf{T}^{(k)}$
	\end{algorithmic}
\end{algorithm}

\subsection{Discussion}
\label{sec:optimization_discuss}
The proposed optimization framework leverages the MM principle to address the nonlinear coupling between the transport plan $\mathbf{T}$ and the GMM parameters $\Theta$ in the joint objective~\eqref{eq:original_objective}. 
Importantly, the MM reformulation induces a \emph{self-adaptive weighting} of different foundation models. Specifically, the coefficients
\[
\lambda_{P,v}^{(k)} = 2\langle\mathbf{T}^{(k)}, \mathbf{P}_v^{(k)}\rangle, 
\quad
\mu_{\hat{P},v}^{(k)} = 2\langle\mathbf{T}^{(k)}, \hat{\mathbf{P}}_v^{(k)}\rangle
\]
are updated at each iteration based on the current transport cost for each model. Models with lower transport cost, indicating stronger semantic alignment, receive larger weights in the subsequent OT step, thereby exerting greater influence on the updated transport plan. This adaptive mechanism eliminates the need for manual weight tuning.
Furthermore, the coupling between $\mathbf{T}$ and $\Theta$ forms a closed-loop refinement: the updated $\mathbf{T}$ yields semantically informed soft assignments that guide the GMM parameter updates, while the refined $\Theta$ produces posterior matrices $\mathbf{P}_v(\Theta)$ that reshape the cost structure in the OT problem. Over iterations, this synergy progressively improves alignment quality across heterogeneous foundation models.
\section{Experimental Setup.}
\label{sec:setup}
\subsection{Datasets.} 
\label{sec:datasets}
The natural datasets include: \texttt{ImageNet}~\cite{imagenet}, \texttt{SUN397}~\cite{sun397}, \texttt{Aircraft}~\cite{aircraft}, \texttt{EuroSat}~\cite{eurosat}, \texttt{StanfordCars}~\cite{scars}, \texttt{Food101}~\cite{food101}, \texttt{Pets}~\cite{pets}, \texttt{Flower102}~\cite{flowers}, \texttt{Caltech101}~\cite{caltech101}, \texttt{DTD}~\cite{dtd}, and \texttt{UCF101}~\cite{ucf101}. 
The remote sensing datasets include: \texttt{AID}~\cite{AID}, \texttt{EuroSAT}~\cite{eurosat}, \texttt{MLRSNet}~\cite{MLRSNet}, \texttt{OPTIMAL31}~\cite{OPTIMAL}, \texttt{PatternNet}~\cite{PatternNet}, \texttt{RESISC45}~\cite{RESISC45}, \texttt{RSC11}~\cite{RSC11}, \texttt{RSICB128}~\cite{RSICB}, \texttt{RSICB256}~\cite{RSICB}, and \texttt{WHURS19}~\cite{WHURS19}. 
The medical datasets include: \texttt{SICAP-MIL}~\cite{sicap}, \texttt{SKINCANCER}~\cite{skincancer}, \texttt{LC-LUNG}~\cite{lung}, \texttt{NCT-CRC}~\cite{nct}, \texttt{WSSS4LUDA}~\cite{luad}.
These datasets encompass a broad spectrum of image classification, allowing us to comprehensively assess the robustness and generalization capability of our method across distinct domains. The specific dataset templates can be found in the code.

\subsection{Implementation details.}
\label{sec:details}
\textbf{Models.}
Our framework is built upon publicly available implementations of both vision-language models (VLMs) and vision foundation models (VFMs). For natural datasets, we employ CLIP ViT-B/16 and CLIP Resnet-50, released by OpenAI, as the primary VLMs without any additional fine-tuning. For extracting visual priors, we adopt DINOv2 ViT-L/14 and DINOv3 ViT-L/16, chosen for their strong clustering capabilities. 
For remote sensing datasets, we use CLIP~\cite{radford2021learning}, RemoteCLIP~\cite{remoteclip}, SkyCLIP~\cite{skyscript} and GeoRSCLIP~\cite{georsclip} as VLMs.
For medical pathology datasets, we use CLIP~\cite{radford2021learning}, {CONCH}~\cite{conchlu2024visual}, {PLIP}~\cite{pliphuang2023visual}, and {MUSK}~\cite{muskxiang2025vision} as VLMs. Additionally, the image encoder from VLMs is reused as an auxiliary visual encoder.

\noindent\textbf{Process.}
During inference, each image is processed using a single center-cropped view of size $224 \times 224$ to reduce computational overhead. Class names are embedded using a fixed prompt template, with no prompt ensembling or optimization. 
We set the temperature parameter $\epsilon = 0.01$ to control entropy regularization. During the initialization phase, we initialize both for $V_1$ and $V_2$ with equal values, subject to the normalization constraint $\sum_{v=1}^{V_1} \lambda_v=1$ and $\sum_{v=1}^{V_2} \mu_v=1$.
In the \textbf{Transductive} setting, our method is applied directly to test data. In the inductive setting, we first execute our algorithm on the training data to learn both the GMM parameters corresponding to different visual models and weight coefficients. The test data are then processed by obtaining the posterior probabilities of GMM and integrating multiple outputs using the learned weight coefficients to produce the final prediction.
All models operate in inference-only mode without gradient updates. Experiments are conducted on a single NVIDIA RTX 4090 GPU with 24GB of memory.

\section{Extended Discussions}
\label{sec:discussion}

\begin{figure}[t]
    \centering
    \begin{subfigure}[b]{0.45\textwidth}
        \centering
        \includegraphics[width=\textwidth]{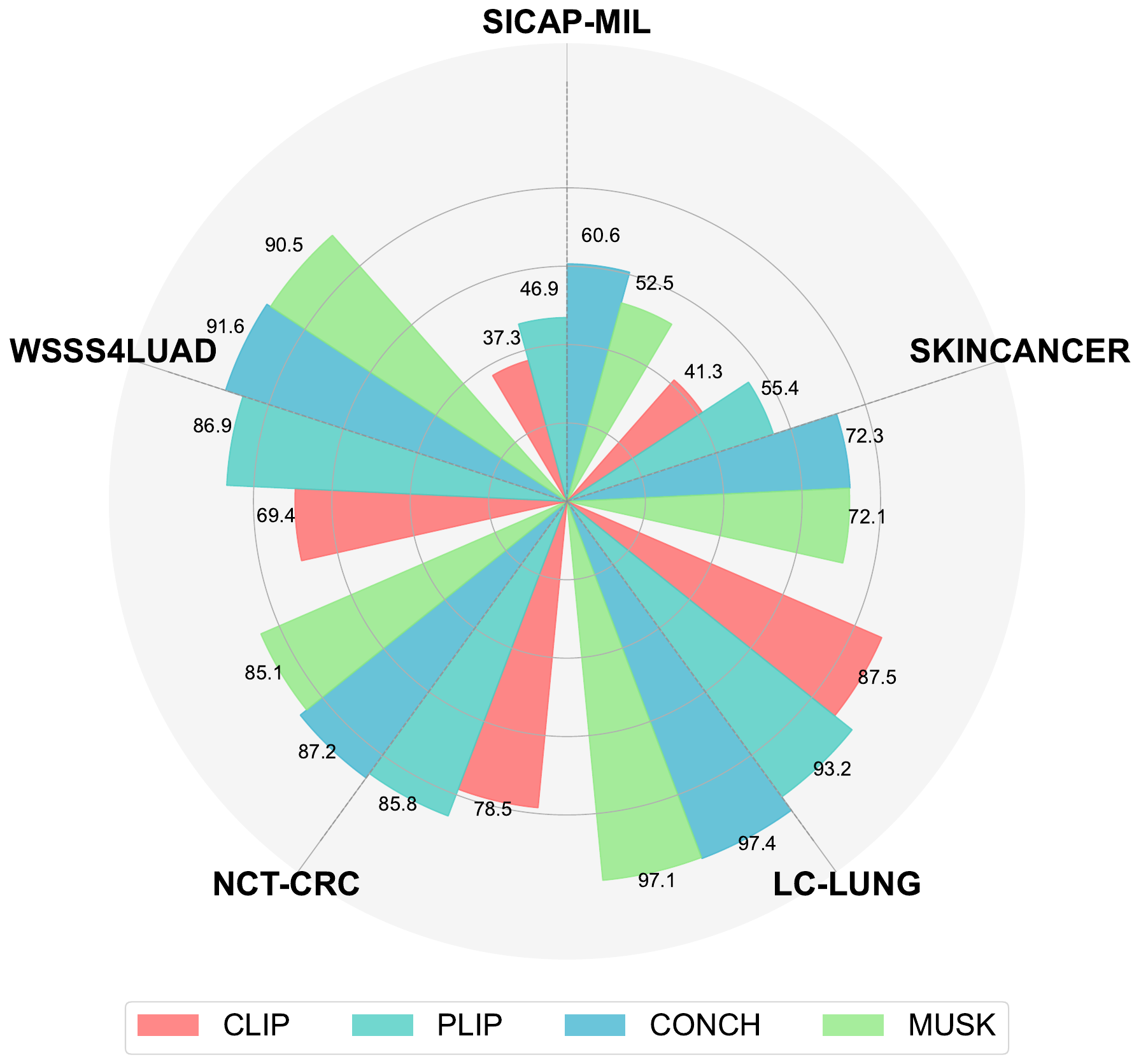}
        \caption{Medical datasets}
        \label{fig:medical_radial}
    \end{subfigure}
    \hfill
    \begin{subfigure}[b]{0.45\textwidth}
        \centering
        \includegraphics[width=\textwidth]{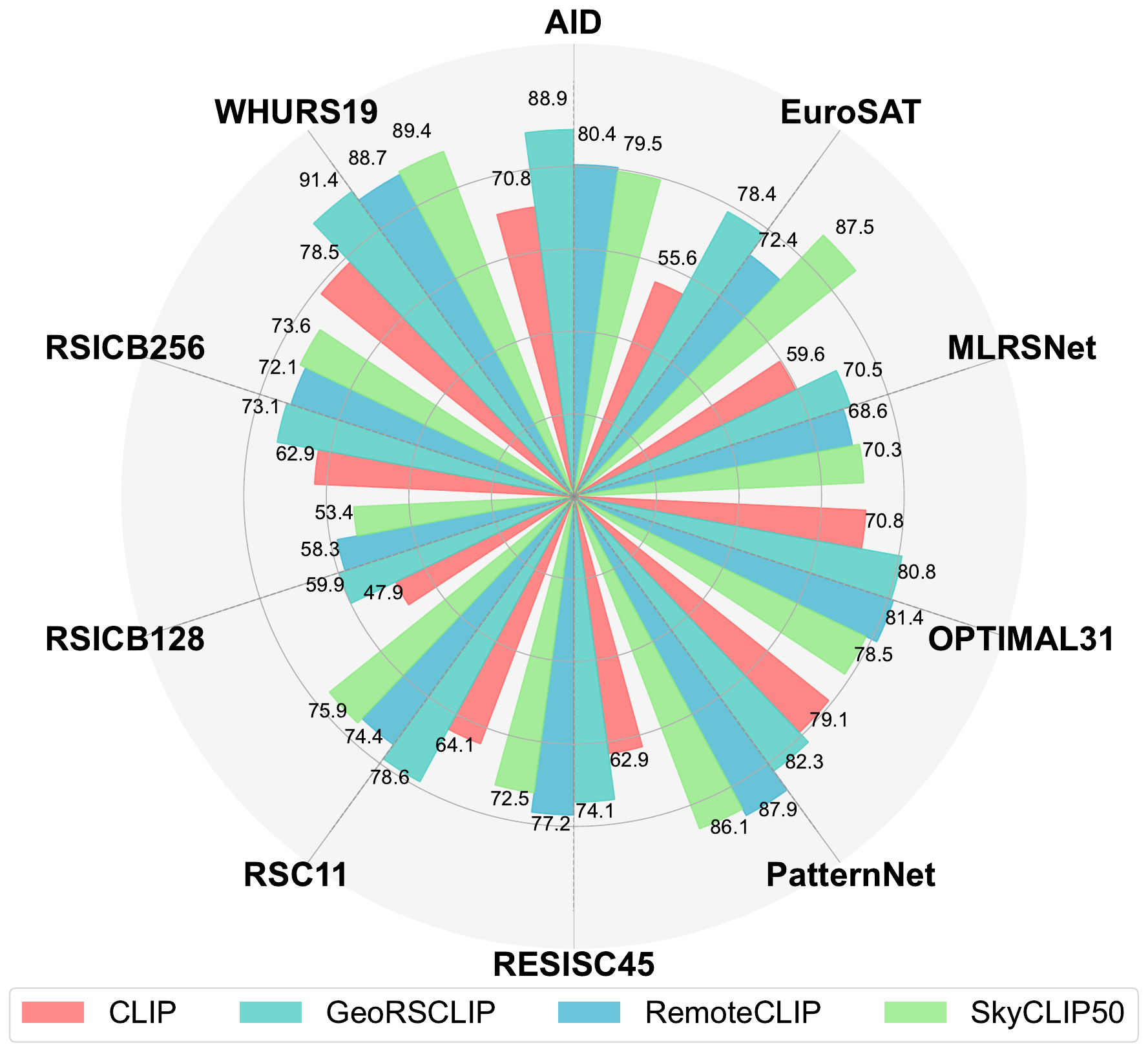}
        \caption{Remote sensing datasets}
        \label{fig:sensing_radial}
    \end{subfigure}
    
    \caption{Clustering accuracy comparison of visual features extracted from different foundation models on medical pathology and remote sensing datasets.}
    \label{fig:radial_bar}
\end{figure}

\subsection{How important is VFM introduction?}
\label{sec:vfms}
To further highlight the importance of visual features, we conduct analyses on remote sensing and medical image datasets by extracting features from the visual branches of different vision-language models. As shown in Fig.~\ref{fig:radial_bar}, the clustering accuracy of visual features is consistently and substantially higher than the corresponding baseline models’ zeroshot performance. In particular, although CONCH achieves the highest clustering accuracy across all datasets, it is nevertheless outperformed by MUSK on the zero-shot task. This observation reveals an inherent modality gap in vision-language models: a naïve strategy that assigns categories solely based on the similarity between sample visual features and candidate class prototypes proves ineffective. Therefore, incorporating high-quality visual features as a complementary source of information is crucial for improving overall performance.

\subsection{Qualitative Visualization and Analysis}
\label{sec:visualization}
\begin{figure*}[h]
	\centering
	\includegraphics[width=1\textwidth]{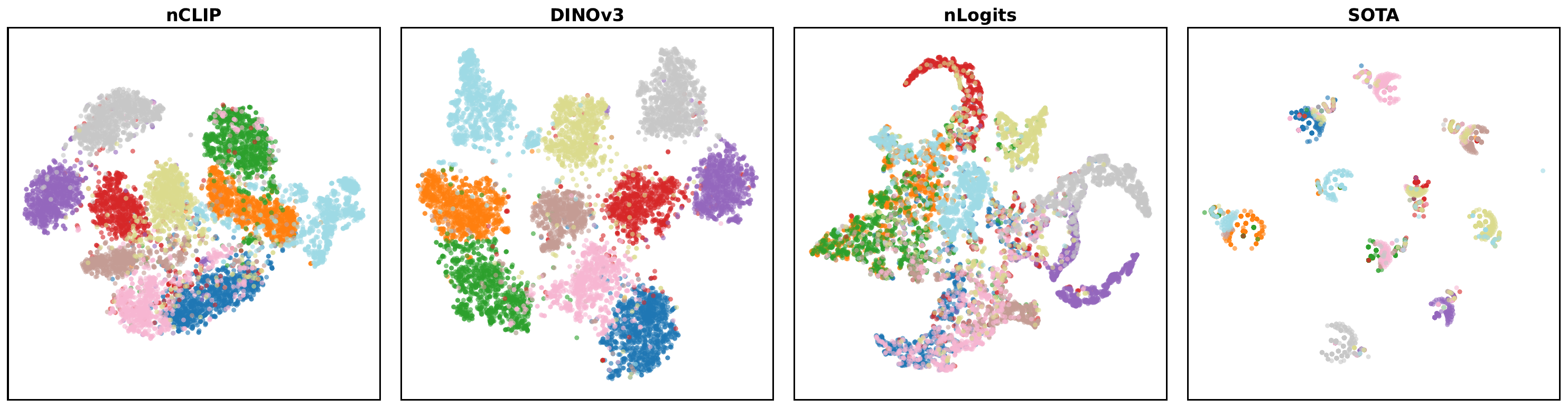}
	\includegraphics[width=1\textwidth]{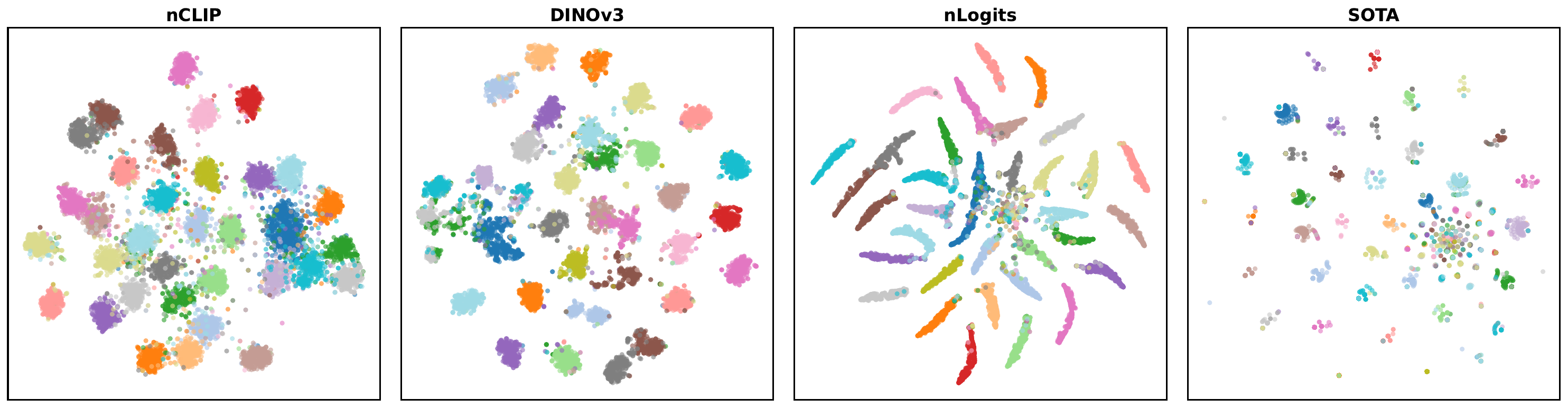}
	
	\caption{t-SNE visualization of predicted clusters on the \texttt{EuroSat}(first row) and \texttt{Food101}(second row) dataset. SOTA significantly improves cluster compactness and separation over CLIP, highlighting superior integration of visual and semantic cues.}
	\label{Fig:tsne}
\end{figure*}

To intuitively compare the differences between visual features from VFMs and outputs from VLMs, we employed t-SNE to visualize output of different models on the \texttt{EuroSat} and \texttt{Food101}. As shown in Fig.~\ref{Fig:tsne}, the experiment included four types of output sources: \textbf{nCLIP} (Visual representations from CLIP visual encoder), \textbf{DINOv3} (Visual representations from DINOv3), \textbf{nLogits} (Outputs from raw CLIP), and \textbf{SOTA} (Outputs from raw CLIP optimized with SOTA). The results reveal that while the visual features (e.g., from CLIP and DINOv3) show clearer and more compact clusters, the zero-shot logits from CLIP remain poorly separated due to the strong modality misalignment between visual and textual spaces. Among the visual encoders, DINOv3 exhibits superior feature organization and stronger inter-class separation than CLIP, highlighting its advantage in capturing discriminative structures. These observations confirm that our adaptive optimal transport alignment effectively exploits the structured visual information from VFMs to calibrate and refine the multimodal logits, thereby enhancing the zero-shot classification performance of VLMs.

\subsection{Computational efficiency analysis}
\label{sec:efficiency}
Our ensemble inference is iterative, alternating between GMM parameter updates and OT. With diagonal covariances, one EM update costs \(O(VNKD)\), where \(V\) is the number of VFMs, \(N\) is the number of target samples, \(K\) is the number of Gaussian components, and \(D\) is the feature dimension. The entropic OT is solved by Sinkhorn with cost \(O(T_{\mathrm{SK}}NK)\), where \(T_{\mathrm{SK}}\) is the number of Sinkhorn iterations. Overall, the per-iteration computational complexity is \(O(VNKD + T_{\mathrm{SK}}NK)\). 
Tab.~\ref{tab:comparison} compares representative methods. Compared with graph-based approaches such as ECALP and TransCLIP, which typically incur higher computation and memory overhead due to graph construction and inference, our method achieves competitive efficiency while delivering stronger accuracy. In contrast, ADAPT is more efficient but exhibits inferior accuracy than ours, highlighting that our design offers a favorable balance between effectiveness and efficiency.

\begin{table*}[h]
\centering
\caption{Comparison on remote sensing and medical pathology (mean over datasets). ``max''/``mean'' are ensemble rules; adaptive methods (e.g., TransCLIP and ADAPT) are applied per vanilla model before ensembling.}
\label{tab:comparison}
\begin{tabular}{lccc ccc}
\toprule
\multirow{2}{*}{Method} &
\multicolumn{3}{c}{Remote sensing} &
\multicolumn{3}{c}{Medical Pathology} \\
\cmidrule(lr){2-4}\cmidrule(lr){5-7}
& ACC/\% & Time/s & Memory/MiB & ACC/\% & Time/s & Memory/MiB \\
\midrule
Vanilla(max)  & 63.50 & -- & -- & 65.32 & -- & -- \\
Vanilla(mean) & 69.80 & -- & -- & 69.52 & -- & -- \\
ECALP(\scalebox{0.7}{ICLR'25})          & 72.69 & 260.2535 & 3330 & 69.82 & 72.9642 & 1012 \\
TransCLIP(\scalebox{0.7}{NeurIPS'24})       & 76.99 & 1.7839 & 911 & 75.29 & 0.7548 & 1067 \\
ADAPT(\scalebox{0.7}{NeurIPS'25})           & 77.46 & 0.2587 & 1089 & 78.07 & 0.2132 & 970 \\
Ours           & 81.45 & 0.9488 & 986 & 83.9 & 0.3120 & 1435 \\
\bottomrule
\end{tabular}
\end{table*}

\subsection{Model subset analysis}
\label{sec:subset}

For remote-sensing and medical datasets, we further conduct model-subset experiments (Tab.~\ref{tab:subset}). Notably, although CLIP exhibits the weakest cross-domain performance among all VLMs, integrating it with any target-domain-specific model via our framework consistently leads to performance gains, validating the effectiveness of our approach under diverse model subset configurations. This behavior suggests that our method does not rely on a single dominant model; instead, it can leverage diverse model outputs in a cooperative manner. 

\begin{table*}[h]
\centering
\caption{The results of combining different models. The first line is the results of the baseline model.}
\label{tab:subset}
\setlength{\tabcolsep}{6pt}
\renewcommand{\arraystretch}{1.2}

\begin{tabular}{cccccccccc}
\toprule
\multicolumn{5}{c}{Remote Sensing} & \multicolumn{5}{c}{Medical Pathology} \\
\cmidrule(lr){1-5}\cmidrule(lr){6-10}
CLIP & Geo & Remote & Sky & Acc/\% & CLIP & CONCH & MUSK & PLIP & Acc/\% \\
\midrule
56.1 & 64.5 & 61.0 & 64.4 & --    & 30.8 & 62.9 & 58.2 & 66.3 & --    \\
$\checkmark$ & $\checkmark$ &              &              & 76.05 & $\checkmark$ & $\checkmark$ &              &              & 68.25 \\
$\checkmark$ &              & $\checkmark$ &              & 77.18 & $\checkmark$ &              & $\checkmark$ &              & 73.06 \\
$\checkmark$ &              &              & $\checkmark$ & 75.52 & $\checkmark$ &              &              & $\checkmark$ & 68.58 \\
             & $\checkmark$ &              & $\checkmark$ & 80.41 &              & $\checkmark$ & $\checkmark$ &              & 83.07 \\
             & $\checkmark$ &              & $\checkmark$ & 77.58 &              & $\checkmark$ &              & $\checkmark$ & 80.01 \\
             &              & $\checkmark$ & $\checkmark$ & 81.16 &              &              & $\checkmark$ & $\checkmark$ & 81.75 \\
$\checkmark$ & $\checkmark$ & $\checkmark$ &              & 78.90 & $\checkmark$ & $\checkmark$ & $\checkmark$ &              & 79.87 \\
$\checkmark$ & $\checkmark$ &              & $\checkmark$ & 77.09 & $\checkmark$ & $\checkmark$ &              & $\checkmark$ & 79.32 \\
             & $\checkmark$ & $\checkmark$ & $\checkmark$ & 82.40 &              & $\checkmark$ & $\checkmark$ & $\checkmark$ & 84.30 \\
$\checkmark$ & $\checkmark$ & $\checkmark$ & $\checkmark$ & 81.50 & $\checkmark$ & $\checkmark$ & $\checkmark$ & $\checkmark$ & 83.90 \\
\bottomrule
\end{tabular}
\end{table*}
{
    \small
    \bibliographystyle{ieeenat_fullname}
    \bibliography{main}
}


\end{document}